\pdfoutput=1
\pdftrailerid{}
\input{glyphtounicode}
\documentclass[11pt,a4paper]{article}

\usepackage[T1]{fontenc}
\usepackage[utf8]{inputenc}
\usepackage{mathptmx}

\usepackage[margin=2.5cm]{geometry}
\usepackage{setspace}
\usepackage{microtype}
\usepackage{amsmath,amssymb}
\usepackage{booktabs}
\usepackage{longtable}
\usepackage{array}
\usepackage{multirow}
\usepackage{threeparttable}
\usepackage{float}
\usepackage{graphicx}
\usepackage[font=small,labelfont=bf,labelsep=period,justification=justified,singlelinecheck=false]{caption}
\usepackage{titlesec}
\usepackage{enumitem}
\usepackage[numbers,sort&compress]{natbib}
\usepackage{xurl}
\usepackage[hidelinks]{hyperref}

\newcommand{\papertitle}{Frame geometry, fascicle extrapolation and leaderboard overfitting in muscle ultrasound}
\hypersetup{pdftitle={\papertitle}, pdfauthor={Soheil Kooklan}, pdfcreator={}, pdfproducer={}, pdfsubject={}, pdfkeywords={}}

\titleformat{\section}{\large\bfseries}{\thesection}{0.6em}{}
\titleformat{\subsection}{\normalsize\bfseries}{\thesubsection}{0.6em}{}
\titleformat{\paragraph}[runin]{\normalsize\itshape}{}{}{}[.]
\titlespacing*{\section}{0pt}{2.0ex plus 0.5ex}{0.9ex}
\titlespacing*{\subsection}{0pt}{1.6ex plus 0.4ex}{0.6ex}


\newcommand{\entry}[1]{\texttt{#1}}
\newcommand{\auditversion}{2}
\newcommand{\col}[1]{\texttt{#1}}
\newcommand{\PA}{\mathrm{PA}}
\newcommand{\FL}{\mathrm{FL}}
\newcommand{\MT}{\mathrm{MT}}

\begin{document}

\begin{center}
{\LARGE\bfseries \papertitle\par}
\vspace{1.2em}
{\large Soheil Kooklan\par}
\vspace{0.5em}
{\small Independent researcher\\
Correspondence: \href{mailto:soheilkooklan@gmail.com}{soheilkooklan@gmail.com}\\
ORCID: \url{https://orcid.org/0009-0003-5035-7833}\\
Project page: \url{https://soheilkooklan.github.io}\par}
\end{center}
\vspace{0.5em}

\begin{small}
\noindent\textbf{Abstract}\par\smallskip
\noindent\textbf{Background.} Automated readings of pennation angle (PA), fascicle length (FL) and muscle thickness (MT) from B-mode ultrasound depend on the geometry of the stored frame and on how fascicles that leave the image are extended. Challenge entries add a third source of error, settings chosen on a small public leaderboard. This study measures all three in one entry of the UMUD Challenge, which scored 309 frames from six screen layouts, 103 on a public and 206 on a private leaderboard.
\textbf{Methods.} The pipeline applies layout-specific values of scale and pixel aspect, inferred from screen elements or assumed, resamples each frame to square pixels, segments aponeuroses and fascicle fragments with U-Nets and measures PA, FL and MT geometrically. The submitted entry blended its angle and length with public notebooks and took thickness from public notebooks alone. Because the score is additive, 65 single-change contrasts, drawn from 13 base entries, measured each decision on both splits. The pipeline was compared with the raters of the UMUD expert benchmark after the benchmark was audited for tissue-level overlap with the DL\_Track\_US and challenge training images.
\textbf{Results.} The primary selected entry scored 0.27608 on the private split, the 69th reported position of 690 teams; the pipeline-only file \texttt{b\_c93} scored 0.40716. Reading every frame with square pixels cost 0.029 public and 0.037 private; the effect replicated for the Lumify layout and for the other layouts together, not for one Telemed layout. Of 20 late-stage corrections, global and layout-specific, 9 kept the sign of their effect on the private split, and their RMS-based dispersion scale, 0.005 to 0.008, exceeded most of their public effects. The audit found that 26 of 35 benchmark frames share their tissue with training images of both DL\_Track\_US and the challenge. On the 9 frames with no detected match in either set or orientation, the recipe scored 0.213, two experts against two others 0.218 and DL\_Track\_US 0.263 (0.209, 0.196 and 0.259 when both raters of a reference pair are required), with no difference established.
\textbf{Conclusions.} Frame geometry and the extension rule moved readings as much as the choice of model, and small corrections tuned on about 100 public frames did not carry over. Only nine benchmark frames, all of one screen layout, are free of detected tissue-level training overlap, too few to rank methods.

\end{small}
\vspace{0.5em}

\section{Introduction}
\label{sec:intro}

The length of a muscle's fascicles and their angle to the line of pull set much of what the muscle can do: how far it can shorten, how fast, and how much of the force of its fibres reaches the tendon \citep{lieber2000}. In living people these quantities are measured mostly with B-mode ultrasound, where a longitudinal frame shows the superficial and deep aponeuroses with the fascicles between them \citep{franchi2018}. Manual readings are reliable \citep{kwah2013} but slow, and automated methods have been developed to replace them (Section~\ref{sec:related}).

Two sources of error in automated reading lie outside segmentation. The first is the geometry of the stored frame. Exported frames differ in millimetre scale, and on some of them lateral and vertical pixels do not span the same distance; reading such a frame as square leaves vertical distances unchanged but alters every angle and every oblique length. The second is extrapolation. In long muscles a fascicle rarely fits in the field of view, and because $\FL = \MT/\sin\PA$ for a straight fascicle between parallel aponeuroses, a small angle error becomes a large length error when fascicles are flat. Public challenges add a statistical problem: a leaderboard that scores a fixed public subset can be reused until its scores describe the subset more than the task, which is called adaptive overfitting \citep{blum2015,dwork2015}. Its size is rarely measured in challenge reports.

The UMUD Challenge on Kaggle asked for pennation angle (PA), fascicle length (FL) and muscle thickness (MT) on 309 test frames, scored on a public split of 103 frames and a private split of 206 \citep{umudkaggle}. I took part as an individual participant. This study uses that entry as a case study of the three sources of variation. It quantifies the effect of a layout-derived geometric correction of the frames, compares conventions for extending fascicles beyond the image, and measures how often improvements found on the public split held on the private split. A tissue-level audit of the public expert benchmark then shows how far that benchmark can serve as independent evidence. Public notebooks contributed to the submitted entry, and their part is kept separate from the pipeline throughout. The study analyses one challenge entry; it does not propose a new network architecture or a clinically validated measurement system.

\section{Related work}
\label{sec:related}

\paragraph{Tracking and image-based methods} Many tools follow fascicles through a sequence from a user-defined start, by optical flow \citep{cronin2011,farris2016}, by frame-wise estimation \citep{vanderzee2022} or by both with a Kalman filter \citep{vanderzee2025}. Others estimate orientation from image texture, with a Hough transform \citep{zhou2008}, a tube filter followed by a Radon transform \citep{rana2009} or spatial and frequency filtering in the SMA macro \citep{seynnes2020}. For length, the open question is how to extend a fascicle beyond the image: in the biceps femoris, linear extrapolation from static images overestimated length by 8 to 11\% against extended field-of-view scans \citep{pimenta2018}, and trigonometric estimates varied more between subjects than other single-image methods \citep{brusco2022}.

\paragraph{Deep learning} U-Net models now segment aponeuroses and fascicles in single frames. DL\_\allowbreak Track\_\allowbreak US computes the three quantities from such masks \citep{cronin2020,ritsche2023joss} and differed from manual analysis by $-2.4$~mm in mean fascicle length, less than 1.5\textdegree{} in angle and less than 1~mm in thickness \citep{ritsche2024}. In an Attention U-Net pipeline, length errors grew where fascicles left the space between the aponeuroses \citep{katakis2022}. The UMUD web application collects open musculoskeletal ultrasound datasets, including an expert-analysed benchmark of 35 architecture frames \citep{ritsche2026,umudosf}.

\paragraph{Leaderboards} Repeated use of one holdout set to select models makes its scores optimistic, and leaderboard designs that limit the information each submission returns have been proposed \citep{blum2015,dwork2015}. These analyses are general and give no measured size for a real challenge.

\paragraph{Relation to this study} Most evaluations above compared automated and manual readings on frames with a known scale and under one extension rule, and the papers cited do not report a check of their test images against the tools' training data. This study measures what those evaluations held fixed: the effect of the stored frame's geometry on readings and scores across six screen layouts, the spread between extension rules on the same frames, the share of public-split corrections that held on a private split, and the overlap of a public benchmark with a published training set.

\section{Materials and methods}
\label{sec:methods}

\subsection{Challenge data and score}
\label{sec:data}

The training data are 2761 frames with fascicle masks and 1048 frames with aponeurosis masks from earlier publications, without angle, length or thickness labels \citep{umudkaggle}. The 309 test frames come from a Siemens Acuson Juniper, a Telemed ArtUS EXT-1H and a Philips Lumify, show the vastus lateralis, gastrocnemius medialis and rectus femoris, and include participants with cerebral palsy who do not appear in the training set. Two researchers annotated every test frame with all three readings, and their means are the labels, so each split holds 103 or 206 frames for every target. Which frames form the public and the private split is not published. The organiser's scorer \citep{umudscore} weights every frame of a split equally and returns
\begin{equation}
\begin{aligned}
S &= S_1 + 10^{-6}\,S_2 + 10^{-9}\,S_3,\\
S_1 &= \frac{1}{3}\left(\frac{\mathrm{MAE}_{\PA}}{6^\circ} + \frac{\mathrm{MAE}_{\FL}}{12~\mathrm{mm}} + \frac{\mathrm{MAE}_{\MT}}{3~\mathrm{mm}}\right),
\end{aligned}
\label{eq:score}
\end{equation}
where $S_2$ and $S_3$ repeat $S_1$ with the median absolute error and the root mean square error. The scorer describes the three tolerances as minimal detectable changes from the literature. $S_2$ and $S_3$ only break ties: since the median absolute error is close to the mean absolute error here, $10^{-6}S_2$ is about 2 to $5 \times 10^{-7}$, enough to move a score shown to five decimals by at most one unit in the last place. Sums and differences of displayed scores carry the same rounding, up to one unit in the fifth decimal. In $S_1$, one millimetre of thickness error costs as much as four millimetres of length or two degrees of angle. The definitions of the readings differ between sources (Table~\ref{tab:defs}). The pipeline's thickness is therefore called the vertical inter-aponeurosis gap, and the perpendicular gap is computed beside it.

\begin{table}[ht]
\centering
\caption{Operational definitions of the three readings.}
\label{tab:defs}
\small
\begin{tabular}{l >{\raggedright\arraybackslash}p{4.3cm} >{\raggedright\arraybackslash}p{3.5cm} >{\raggedright\arraybackslash}p{4.6cm}}
\toprule
Reading & Challenge (overview; forum) & Benchmark (readme) & Pipeline \\
\midrule
MT & overview: perpendicular distance; later forum answer: three vertical lines between inner borders \citep{umud689019} & three straight lines, direction not stated & vertical inner gap at 20, 50, 80\% of the shared width; perpendicular gap for comparison \\
FL & three fascicles needing the least extrapolation, extended linearly & three segmented lines & recipe, Eq.~\eqref{eq:trust}; \col{fl\_lin3} and \col{fl\_centre} \\
PA & angle to the deep aponeurosis, three angles & three angles & length-weighted median fragment angle \\
Scale & ruler on the frame & given per frame & from the screen layout (Table~\ref{tab:layouts}) \\
\bottomrule
\end{tabular}
\end{table}

\subsection{Layout-derived geometric correction}
\label{sec:layouts}

Each test frame is a screenshot of a scanner display. A layout module assigned every frame to one of six screen layouts by its size, file type and fixed interface pixels and read the vertical scale $v$ from the depth ruler or from fixed ruler spans (Table~\ref{tab:layouts}). The pixel aspect $a$, the ratio of lateral to vertical pixels per millimetre in the stored frame, was inferred from screen elements: the display size on the Siemens layout, the tick spacings of the two rulers on the Telemed layouts and the shape of the round probe mark on the Lumify layout. On the grey and crop layouts $a = 1$ was assumed. The layout rules and aspect values follow public analyses \citep{ambrosm2026,watehell2026}; this work implemented them in one module and measured their effect. None was checked against scanner metadata or a phantom, so the correction is layout-derived and is not a physical calibration.

\begin{table}[ht]
\centering
\caption{Screen layouts of the test frames, vertical scale, and pixel aspect $a$ with its source. Values marked as assumed carry no evidence.}
\label{tab:layouts}
\small
\begin{tabular}{l r >{\raggedright\arraybackslash}p{3.7cm} r >{\raggedright\arraybackslash}p{4.4cm}}
\toprule
Layout & Frames & Vertical scale & $a$ & Source of $a$ \\
\midrule
Siemens, $800 \times 1200$ & 91 & depth ruler, median 15.23 px/mm & 1.125 & inferred display resize \\
Telemed, $800 \times 1200$ & 90 & fixed span, 14.82 px/mm & 0.908 & ruler tick spacings \\
Philips Lumify & 58 & depth ruler, median 15.0 px/mm & 0.836 & probe mark taken as a circle \\
Telemed, $644 \times 1088$ & 50 & fixed span, 12.61 px/mm & 1.001 & ruler tick spacings \\
Grey interface & 12 & bottom ticks as 1~cm, 16.7 px/mm & 1.0 & assumed \\
Crops & 8 & bottom ticks as 1~cm, 7.78 px/mm & 1.0 & assumed \\
\bottomrule
\end{tabular}
\end{table}

\begin{figure}[ht]
\centering
\includegraphics[width=\linewidth]{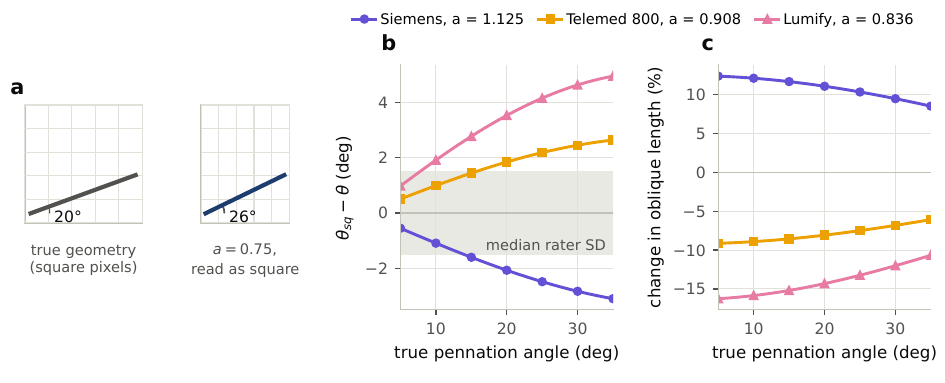}
\caption{Frame geometry. (a)~A fascicle at 20\textdegree{} and the same frame stored with $a = 0.75$ and read as square. (b)~Angle read on square pixels minus the true angle for the three layouts with $a \neq 1$, from Eq.~\eqref{eq:aspect}; the band is $\pm$ the median standard deviation between benchmark raters. (c)~Change of an oblique length.}
\label{fig:framegeom}
\end{figure}

A segment spanning $\Delta x$ lateral and $\Delta y$ vertical pixels spans $\Delta x/(av)$ and $\Delta y/v$ millimetres, so its angle $\theta$ to the horizontal satisfies
\begin{equation}
\tan\theta = a \tan\theta_{\mathrm{px}}.
\label{eq:aspect}
\end{equation}
The networks segment the stored crop of each frame at their own input size ($384 \times 576$ pixels for aponeuroses, $512 \times 512$ for fascicles); their probability maps are resized back to the crop and then resampled laterally to a width of $\mathrm{round}(w_{\mathrm{crop}}/a)$ pixels, and all measurement, including the 25-pixel fragment limit below, runs on this isotropic grid at the native vertical scale of the test frames (7.8 to 16.7 px/mm, so the limit spans 1.5 to 3.2~mm). Reading a frame as square means applying the vertical scale $v$ along both axes. Read as square, a segment at $\theta$ to the horizontal appears at $\theta_{\mathrm{sq}} = \arctan(\tan\theta/a)$ and its length becomes $L_{\mathrm{sq}} = L(\sin^2\theta + a^2\cos^2\theta)^{1/2}$, while vertical distances do not change (Figure~\ref{fig:framegeom}). These relations take the deep aponeurosis as horizontal. For a deep aponeurosis inclined at $\beta$ to the horizontal, measured in the same rotational sense as the fascicle so that the fascicle lies at $\theta = \PA + \beta$, the fascicle and the aponeurosis are distorted separately, $\PA_{\mathrm{sq}} = \arctan\{\tan(\PA+\beta)/a\} - \arctan(\tan\beta/a)$, and the closed form below, which keeps the vertical gap, scales by $\sin\PA/\sin\PA_{\mathrm{sq}}$ and not by $L_{\mathrm{sq}}/L$. The sensitivity of the readings to an error in $a$ or $v$, systematic within a layout, was estimated by Monte Carlo on the pipeline's test readings without labels (Supplementary Section~\ref{app:mc}).

\subsection{Segmentation and measurement}
\label{sec:seg}
\label{sec:geometry}

Aponeuroses were segmented by a U-Net \citep{ronneberger2015} with an ImageNet-initialised ResNet-34 encoder \citep{he2016,iakubovskii2019}, trained in five folds on the challenge masks; the training notebook stops unless the ImageNet encoder loads. Folds grouped near-identical frames by difference hash, since participant and recording identifiers are not available. Each fold's threshold was matched to the thickness of the annotated masks, and inference averages the five folds with flips. Fascicles were segmented by a single-fold U-Net of the same design, trained with an earlier code version (0.12.0). Its log records 814 duplicates removed from the 2761 fascicle frames, leaving 1947 pairs, 1558 for training and 389 for validation, and a best validation Dice of 0.488. The five aponeurosis folds reached a pooled held-out Dice of 0.79 to 0.81, an optimistic figure because epoch and threshold were chosen on the same frames; no hash ties the fascicle checkpoint to its log (Supplementary Tables~\ref{tab:train} and~\ref{tab:segval}). The models are working parts of the pipeline; this study makes no claim about their accuracy, and because folds were not grouped by participant, the results say nothing about generalisation to new participants.

\begin{figure}[ht]
\centering
\includegraphics[width=\linewidth]{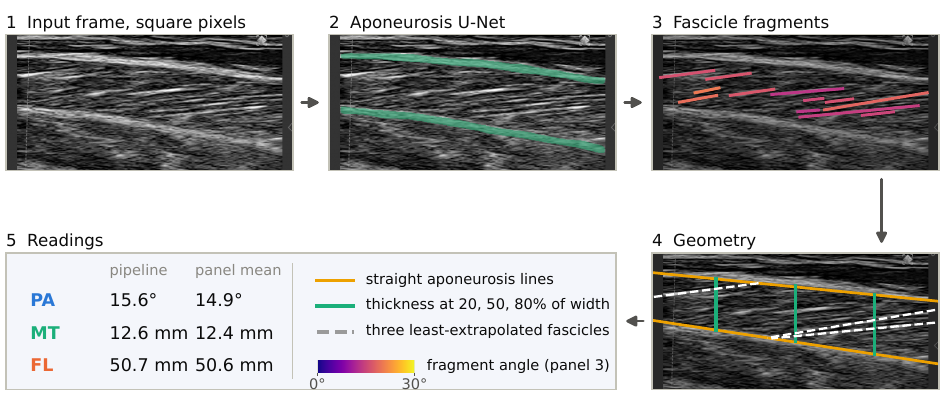}
\caption{The pipeline on benchmark frame im\_11 (OSF, CC BY 4.0). Panel 5 gives the pipeline's readings beside the mean of the benchmark raters. The dashed lines in panel 4 are the three least-extrapolated fascicles of \col{fl\_lin3}, while the length in panel 5 is the recipe of Eq.~\eqref{eq:trust}. This frame shares its tissue with a DL\_Track\_US training image (Section~\ref{sec:res-bench}) and only illustrates the steps.}
\label{fig:workflow}
\end{figure}

\begin{figure}[ht]
\centering
\includegraphics[width=\linewidth]{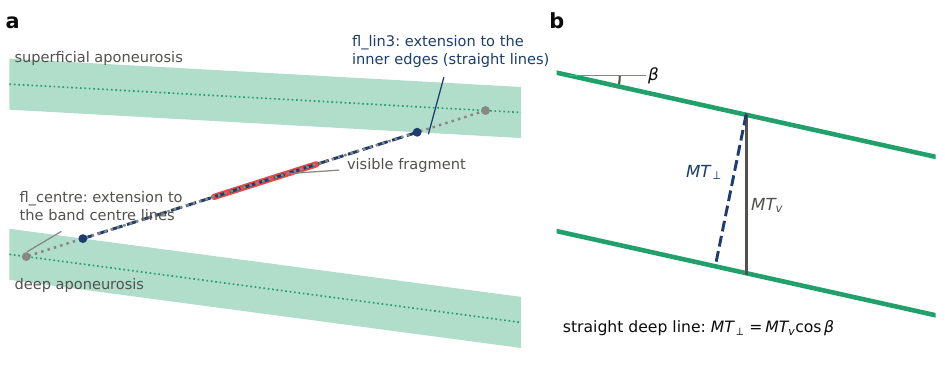}
\caption{Conventions behind the readings. (a)~A visible fragment extended to straight inner edges (\col{fl\_lin3}, dashed) or to the band centre lines (\col{fl\_centre}, dotted). (b)~Vertical and perpendicular thickness on aponeuroses inclined at $\beta$.}
\label{fig:convschem}
\end{figure}

Figure~\ref{fig:workflow} shows the measurement. Each aponeurosis band is fitted with a quadratic along its inner edge. The vertical gap $\MT_v$ (\col{mt\_mm}) is the distance between the inner edges along the image column at 20, 50 and 80\% of the shared width, averaged. The perpendicular gap $\MT_\perp$ is the shortest distance from the superficial curve at the same columns to the deep curve, averaged; when the deep aponeurosis is straight and inclined at $\beta$ it equals $\MT_v\cos\beta$ at every column, whatever the slope of the superficial one (Figure~\ref{fig:convschem}b), and curvature of the deep aponeurosis makes the relation approximate. $\MT_v$ was kept for the challenge recipe because the organiser described vertical lines; $\MT_\perp$ is the distance that a straight fascicle between parallel aponeuroses calls for. The two are reported as distinct operational measurements and not as interchangeable definitions of muscle thickness. Fascicle fragments are line-shaped components of the fascicle mask at least 25 pixels long; the frame's angle is the length-weighted median of their angles to the local deep tangent between 5 and 45\textdegree. The recipe adds $+1.2^\circ$, set from public scores early in the competition, before the length and its weight below are computed.

Two closed forms follow from the two gaps, $\FL_{c,v} = \MT_v/\sin\PA$ and $\FL_{c,\perp} = \MT_\perp/\sin\PA$. The recipe uses $\FL_{c,v}$ and shrinks it towards a fixed length $F_0$,
\begin{equation}
\FL = w\,\FL_{c,v} + (1-w)\,F_0, \qquad w = \frac{1}{1 + (k\cot\PA)^2},
\label{eq:trust}
\end{equation}
with $k = 0.176$ and $F_0 = 93$~mm chosen on the public leaderboard; $\FL_{c,\perp}$ enters only the comparison of Table~\ref{tab:length}. An angle error $\sigma_\theta$ gives the closed form a relative error of about $\sigma_\theta\cot\PA$, so Eq.~\eqref{eq:trust} resembles the posterior mean of a normal prior $N(F_0, \sigma_F^2)$ and a normal measurement, with $k$ in place of $\FL_{c,v}\sigma_\theta/\sigma_F$. The analogy is loose: $k$ is treated as constant although it depends on the angle, the error of the closed form is skewed because $1/\sin\PA$ is convex, the error of thickness is ignored, and a posterior mean minimises squared error while the score uses absolute error. Equation~\eqref{eq:trust} is therefore a heuristic, score-optimised estimator for the challenge labels and not an anatomical measurement of one fascicle, and its weight $w$ is not a calibrated uncertainty. $\FL_{c,v}$ is a vertical-gap surrogate kept to follow the challenge's thickness description; $\FL_{c,\perp}$ is the perpendicular-gap geometric estimate for a straight fascicle between parallel aponeuroses.

Two readings extend fragments along their own line (Figure~\ref{fig:convschem}a). \col{fl\_centre} goes to the band centre lines and takes the length-weighted median, as in Mawra Musawwar's notebook \citep{musawwar2026}. \col{fl\_lin3}, written for this work from the organiser's description of the protocol, extends each fragment to straight lines fitted to the two inner edges and averages the full lengths of the three fragments whose visible part covers the largest share of their length. It approximates a rater's choice of the least-extrapolated fascicles; the raters' choices are not recorded, so the match cannot be checked fascicle by fascicle.

\subsection{Recordings and failed frames}
\label{sec:recordings}

Frames with consecutive file numbers, the same layout and scale and correlated detail thumbnails were grouped into recordings; the data carry no recording identifier, so the grouping is inferred. It gave 169 single frames and 28 recordings of five frames, and every reading of a recording was replaced by its median, which changed 130 of the 140 frames in recordings. Four frames, all single frames, gave no pair of aponeuroses; they received the median of the successful frames of their layout, before the recording median was taken and without a flag in the submitted file. The recording median was not tested on the leaderboard, so its effect on the score is unknown; by the triangle inequality on the saved readings, it changed $S_1$ by at most 0.013 for the pipeline file and by at most 0.009 for the pipeline's share of the blend. On the benchmark each frame formed its own recording and no frame failed. Two internal indicators were tested afterwards as quality flags: fewer than 10 fragments inside the belly, and a difference of more than 3\textdegree{} between the fragment angle and a texture angle.

\subsection{Blend and leaderboard use}
\label{sec:blend}

Three public notebooks were run unchanged and only their prediction files were used: Mawra Musawwar's two-U-Net notebook (M) \citep{musawwar2026}, prvsiyan's notebook (T) \citep{trying2026} and Rishabh Bafna's notebook (K) \citep{bafna2026}. M and K share watehell's U-Net weights \citep{watehell2026w} and measurement code, and T uses public ResNet-34 weights as well. Table~\ref{tab:compose} gives the selected entry. The final thickness owes nothing to the pipeline. In the angle, the pipeline's $+1.2^\circ$ is halved by the average and the blend adds $0.4^\circ$, a net $+1.0^\circ$; the length weights sum to 0.985, which shortens the blend by about 1.5\%.

\begin{table}[ht]
\centering
\caption{Composition of the selected entry \entry{n\_pa\_tm\_up} and the source of each constant. LB: chosen by hand from public scores; rows refer to Supplementary Table~\ref{tab:subs}.}
\label{tab:compose}
\small
\begin{tabular}{l >{\raggedright\arraybackslash}p{5.0cm} >{\raggedright\arraybackslash}p{4.4cm} l}
\toprule
Element & Value & Set by & Without LB \\
\midrule
Angle & $0.5\,\PA_{\mathrm{own}} + 0.5\,\PA_{\mathrm{M}} + 0.4^\circ$ & LB, rows 26--42 & no \\
Length & $0.55\,\FL_{\mathrm{M}} + 0.30\,\FL_{\mathrm{own}} + 0.15 \times 0.9\,\col{fl\_centre}$ & LB, rows 26--45; 0.9 from M & no \\
Thickness & $(\MT_{\mathrm{M}} + \MT_{\mathrm{T}} + \MT_{\mathrm{K}})/3 + 0.1$~mm & LB, rows 32, 60, 65 & no \\
Per-layout terms & lengths $\times 1.03$ Siemens and Lumify, $\times 1.06$ crop and grey; Telemed angles $+0.5^\circ$ & LB, rows 58--70 & no \\
Pipeline angle offset & $+1.2^\circ$ & LB, early entries & no \\
Shrinkage & $k = 0.176$, $F_0 = 93$~mm & LB; $F_0$ started at 83~mm from probes & no \\
Thresholds & 0.955 aponeurosis, 0.9 fascicle & held-out mask thickness & yes \\
Layout geometry & Table~\ref{tab:layouts} & screen elements & yes \\
\bottomrule
\end{tabular}
\end{table}

The main term $S_1$ of Eq.~\eqref{eq:score} is a mean over targets of means over frames, so changes to different columns, or to disjoint frames of one column, add. The blend clips each value to the label range (5 to 45\textdegree, 30 to 200~mm, 10 to 50~mm); no submitted value reached a bound (angles 10.1 to 31.8\textdegree, lengths 31.8 to 134.3~mm, thicknesses 12.4 to 33.8~mm), so additivity holds exactly for $S_1$. Every derived base and every prediction by additivity in this paper replaces whole columns or the rows of whole layouts; none mixes two predictions within one column and frame, where absolute errors would not add. The tie-break terms are also sums over targets and add across columns; within one column they do not add over frames, but they stay below the last displayed decimal. Constants were set by hand from single-change submissions, never for an individual frame or recording. The organiser wrote that weights set by hand after looking at earlier submissions are allowed if the entry runs on new data \citep{umudthread}, that one parameter per scanner type is a global design choice, and that the leaderboard must not tune individual test cases \citep{umud757415,umud758376}. After the close the organisers asked the five top-placed teams for their code \citep{umudwinning}. Two entries were selected for the ranking, \entry{n\_pa\_tm\_up} and \entry{l\_mt\_up1}; Kaggle ranks a team by the better private score of its selected entries, here that of \entry{n\_pa\_tm\_up}. The final private leaderboard, released on 22 November 2026, has one row per team and lists 690 teams, with this one-member team at position 69. No separate decision on its eligibility was received, so this is a reported position.

\subsection{Replication on the private split}
\label{sec:replication}

The private split scored 206 frames that played no part in any decision. Sixty-five contrasts between a single-change submission and its base were collected from 13 base entries (Supplementary Table~\ref{tab:effects}); two bases were derived by additivity. Every frame carries all three targets, so $S_1$ is the mean over frames of $c_i = \tfrac{1}{3}(|e_{\PA,i}|/6^\circ + |e_{\FL,i}|/12~\mathrm{mm} + |e_{\MT,i}|/3~\mathrm{mm})$. Let $d_i$ be the change a submission makes to $c_i$, and $\Delta_{\mathrm{pub}}$ and $\Delta_{\mathrm{priv}}$ its means over the two splits. For new frames of variance $s^2$, $\mathrm{Var}(\Delta_{\mathrm{pub}}) = s^2/103$ and $\mathrm{Var}(\Delta_{\mathrm{pub}} - \Delta_{\mathrm{priv}}) = s^2(1/103 + 1/206)$, so the standard error of a public effect would be $\sqrt{2/3}\,\mathrm{SD}(\Delta_{\mathrm{pub}} - \Delta_{\mathrm{priv}})$. Here the standard deviation is replaced by the root mean square across contrasts, which gives the dispersion scale
\begin{equation}
D_{\mathrm{new}} = \sqrt{2/3}\;\mathrm{RMS}(\Delta_{\mathrm{pub}} - \Delta_{\mathrm{priv}}).
\label{eq:se}
\end{equation}
For the fixed 309 frames split into 103 and 206, the mean over all frames is $\bar d = (103\,\Delta_{\mathrm{pub}} + 206\,\Delta_{\mathrm{priv}})/309$, so $\Delta_{\mathrm{pub}} - \bar d = \tfrac{2}{3}(\Delta_{\mathrm{pub}} - \Delta_{\mathrm{priv}})$ exactly and $D_{\mathrm{fixed}} = \tfrac{2}{3}\,\mathrm{RMS}(\Delta_{\mathrm{pub}} - \Delta_{\mathrm{priv}})$; equivalently, the finite-population factor $1 - 103/309$ on the variance contributes a second factor of $\sqrt{2/3}$. The root mean square pools changes of different kinds: true effects that differ between splits, a mean shift from choosing changes on their public effect, changes of different size, and changes that touch only some layouts. The spread of a single contrast grows with the size of the change and with the share of frames it touches, and frames of one recording are not independent. $D_{\mathrm{new}}$ and $D_{\mathrm{fixed}}$ are therefore dispersion scales for this entry, not conventional standard errors and not a property of the leaderboard. Correlation and sign agreement were resampled by a cluster bootstrap over the 13 bases, and the effect of recordings that span both splits was simulated (Supplementary Table~\ref{tab:splitsim}).

\subsection{Expert benchmark and audit}
\label{sec:benchmethods}

The UMUD benchmark holds 35 frames of the gastrocnemius medialis, soleus and vastus lateralis from four scanners, read by up to seven raters, with the readings of DL\_Track\_US and SMA \citep{umudosf,ritsche2026}. Its scale is given. Rater entries more than 50\% from the median of the other raters for that frame and target were treated as slips and removed (two entries). A reading was scored against the mean of every pair of raters: for each pair and target, the mean absolute error over the frames where both the reading and the pair mean exist was divided by the tolerance, these values were averaged over the 21 pairs without weighting, and the three targets were then averaged as in $S_1$. The human references score one rater against every pair of the others, and every rater pair against every disjoint pair. Each of 2000 resamples (seed 2026) draws as many frames as the set holds, with replacement, and 5 of the 7 raters without replacement, and recomputes all scores on the same draw; differences between readings are taken within a draw, so shared raters and frames enter their intervals. The intervals are percentiles of the resampled scores and need not centre on the seven-rater estimate. The seven raters are treated as a fixed panel, and the draw of 5 of 7, which keeps at least two disjoint pairs, only perturbs the panel; it understates the variance a new panel of raters would bring. A pair mean falls back to the one rater present where the other did not read a frame (R5 read 17 frames). A single reading as reference raises the error of the human references more than that of a fixed reading, so every score was also computed with strict pairs that need both raters (Supplementary Section~\ref{app:nearcopy}). Details and the effect of the slips and of rater R5 are given in Supplementary Section~\ref{app:nearcopy}. A leaderboard-free configuration (\col{pa\_deg} without offset, \col{fl\_lin3}, \col{mt\_mm}) was identified after the results were seen and is post hoc. Each benchmark frame was compared with the 3553 DL\_Track\_US training images by thumbnail correlation (the thumbnail screen, limit 0.97), difference hash, SSIM and keypoint matching restricted to the central tissue region, because interface text differed between two versions of one frame. For the last step both images were resized to 512 pixels wide, the benchmark frame was cut to its central half (rows and columns 25 to 75\%), up to 2000 ORB keypoints were matched by Hamming distance with a ratio test of 0.75, and the inliers of a RANSAC homography (reprojection limit 4 pixels) were counted against the 20 best candidates by thumbnail correlation and by SSIM. The same four measures compared the benchmark with the challenge training images in a Kaggle notebook. The benchmark stores some frames mirrored left to right and others not, and none of these measures is mirror invariant, so the frames left without a match were also matched, as stored and mirrored, against every training image of both sets without the candidate preselection. The challenge audit used every image in the two training folders of the challenge, which hold the 2761 fascicle and 1048 aponeurosis frames, with the same resizing, central crop, ORB and RANSAC settings and candidate rule; images present in both folders were not removed, since a repeated image cannot change a maximum count.

\section{Results}
\label{sec:results}

Leaderboard and benchmark scores come from different frames, raters and scanners, and the benchmark gives the scale; they are not interchangeable. Private effects are taken as the evidence of effect size and public effects as exploratory.

\subsection{Main entries}
\label{sec:res-final}

The primary selected entry, \entry{n\_pa\_tm\_up}, scored 0.23922 public and 0.27608 private (Table~\ref{tab:entries}); the second selected entry scored 0.29184 private, and the leading private score was 0.22681. The pipeline-only file \entry{b\_c93} scored 0.40716 private and notebook M 0.34042. Every listed entry scored worse on the private split: by 0.00029 for \entry{y\_full} and by 0.027 to 0.048 for the others.

\begin{table}[ht]
\centering
\caption{Main entries with public (103 frames) and private (206 frames) scores. Selected entries were submitted on time; late entries were scored after the deadline and are outside the ranking.}
\label{tab:entries}
\small
\begin{tabular}{l l >{\raggedright\arraybackslash}p{6.6cm} r r}
\toprule
Entry & Status & Content & Public & Private \\
\midrule
\entry{y\_full} & on time & pipeline, layout geometry, closed-form length & 0.44484 & 0.44513 \\
\entry{b\_c93} & on time & pipeline only, recipe length; best file of the pipeline & 0.37808 & 0.40716 \\
notebook M & public & run unchanged; private score derived & 0.30652 & 0.34042 \\
\entry{h\_best} & late & blend of Table~\ref{tab:compose} without per-layout terms and without the thickness offset & 0.24843 & 0.27583 \\
\entry{n\_pa\_tm\_up} & selected & \entry{h\_best} with per-layout terms and thickness $+0.1$~mm; primary entry, position 69 & 0.23922 & 0.27608 \\
\entry{l\_mt\_up1} & selected & \entry{h\_best}, lengths $\times 1.03$, thickness $+0.1$~mm & 0.24372 & 0.29184 \\
\entry{p\_safe} & late & one global length scale and thickness offset & 0.24214 & 0.28086 \\
\bottomrule
\end{tabular}
\end{table}

\subsection{Frame geometry}
\label{sec:res-aspect}

Replacing the earlier ruler reader by the layout module moved the best scores from 0.52995 to 0.44484 public and from 0.57348 to 0.44513 private. Reading every frame as square cost 0.02888 public and 0.03688 private. Reading only the Telemed 800 frames as square cost 0.01277 public and gained 0.00152 private, so the other layouts carried 0.01611 and 0.03840. Since the Telemed 644, grey and crop layouts have $a$ close to or set to 1, this remainder is in effect the Siemens and Lumify frames together; the Siemens layout was not tested alone. On the blend, square-pixel angles on the Lumify frames cost 0.02054 public and 0.01693 private, and square-pixel lengths 0.02637 and 0.01778. The correction therefore replicated on new frames for the Lumify layout and for the Siemens and Lumify frames together, and not for the Telemed 800 layout. The Telemed 800 test was made on \entry{y\_full}, whose length is the unshrunk closed form; on these flat fascicles the correction lowers the angle by 1.16\textdegree{} and lengthens $\MT_v/\sin\PA$ markedly, so this test does not separate an error in $a = 0.908$ from the instability of the closed form. On the readings of the test-set run, reading the Telemed 800 frames as square shortens $\MT_v/\sin\PA$ by 9.6~mm on average (from a median of 105~mm) and the recipe length by 5.0~mm, and it moves the closed form from a median of 13.1~mm above notebook M's lengths to 5.4~mm above them. The square reading thus moved the Telemed 800 closed form towards notebook M's lengths, so this contrast may measure the closed form more than the value of $a$. In the Monte Carlo analysis an aspect error of 5\% moved the angle by 0.46 to 0.68\textdegree, below the raters' median spread of 1.52\textdegree, while a vertical scale error of 2\% moved thickness by 0.28 to 0.40~mm, above their spread of 0.26~mm (Supplementary Figure~\ref{fig:calmc}).

\subsection{Fascicle length}
\label{sec:res-length}

\begin{figure}[ht]
\centering
\includegraphics[width=\linewidth]{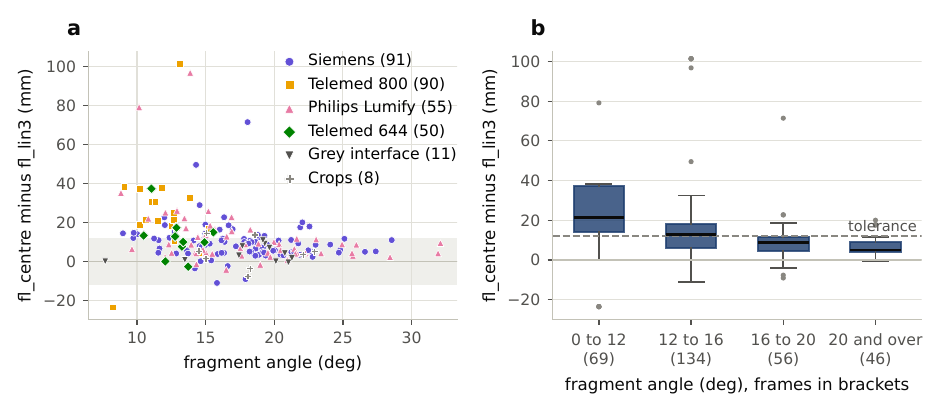}
\caption{Difference between the two extension conventions on the 305 test frames with a pair of aponeuroses (the four failed frames are excluded), from the pipeline's readings without labels. (a)~\col{fl\_centre} minus \col{fl\_lin3} against the fragment angle, by layout; the band is $\pm 12$~mm. (b)~The same difference in bands of fragment angle.}
\label{fig:conventions}
\end{figure}

The extension conventions part company as fascicles flatten (Figure~\ref{fig:conventions}). \col{fl\_centre} exceeded \col{fl\_lin3} by a median of 21.4~mm below 12\textdegree{} and 5.1~mm above 20\textdegree, and by more than the 12~mm tolerance on 48\% of the frames; 45 of the 69 frames below 12\textdegree{} are Telemed 800 frames. In the blend, \col{fl\_lin3} in place of \col{fl\_centre} lost on the public split ($+0.01276$) and gained on the private split ($-0.00146$ at weight 0.15, $-0.00331$ at 0.30), while more weight on \col{fl\_centre} lost on both, more on the private split ($+0.01123$ and $+0.03351$). The larger split and the benchmark raters agree that the least-extrapolated reading is closer to the labels.

On the benchmark (Table~\ref{tab:length}) the closed form $\FL_{c,v}$ read 6.7~mm short on the 25 screen-negative frames; the shrinkage cut its error from 7.32 to 4.90~mm, \col{fl\_lin3} reached 4.32~mm and DL\_Track\_US 3.74~mm, but 16 of these frames are tissue matches with the training images of both DL\_Track\_US and the challenge. On the 9 frames with no detected match the order changed: the recipe reached 2.83~mm, DL\_Track\_US 3.23~mm and \col{fl\_lin3} 3.92~mm. $\FL_{c,\perp}$ was slightly worse than $\FL_{c,v}$, and the recipe with $\MT_\perp$ slightly better than the submitted recipe with $\MT_v$. On the test frames the deep aponeurosis was inclined by a median of 2.8\textdegree, so the perpendicular gap was a median 0.15\% shorter than the vertical gap (about 0.03~mm) and would change the recipe length by 0.08~mm; this was measured on the 295 frames whose saved overlays allowed both aponeuroses to be refitted. The recipe's error fell with panel length, as a pull towards 93~mm, a value above every panel length, predicts (Supplementary Figure~\ref{fig:lenfail}).

\begin{table}[ht]
\centering
\caption{Length readings against the benchmark panel mean on the 25 thumbnail-screen-negative frames, 16 of which are tissue matches, and on the 9 frames with no detected match. Sheet: readings given in the benchmark results sheet. The constant at the panel median of each set uses the labels and serves only as an oracle reference.}
\label{tab:length}
\small
\begin{tabular}{l r r r r}
\toprule
& \multicolumn{2}{c}{Screen negative ($n = 25$)} & \multicolumn{2}{c}{No detected match ($n = 9$)} \\
\cmidrule(lr){2-3}\cmidrule(lr){4-5}
Reading & MAE (mm) & Bias (mm) & MAE (mm) & Bias (mm) \\
\midrule
recipe, Eq.~\eqref{eq:trust}, with $\MT_v$ (submitted) & 4.90 & $+0.99$ & 2.83 & $+2.47$ \\
recipe, Eq.~\eqref{eq:trust}, with $\MT_\perp$ & 4.87 & $+0.58$ & 2.58 & $+2.05$ \\
vertical-gap surrogate $\FL_{c,v} = \MT_v/\sin\PA$ & 7.32 & $-6.71$ & 5.56 & $-5.56$ \\
perpendicular-gap estimate $\FL_{c,\perp} = \MT_\perp/\sin\PA$ & 7.75 & $-7.22$ & 6.08 & $-6.08$ \\
\col{fl\_lin3} (least-extrapolated) & 4.32 & $+2.62$ & 3.92 & $+2.37$ \\
\col{fl\_centre} (band centre lines) & 8.65 & $+8.65$ & 7.98 & $+7.98$ \\
DL\_Track\_US (sheet) & 3.74 & $+0.36$ & 3.23 & $-0.09$ \\
oracle constant at the panel median & 8.67 & $-0.12$ & 6.75 & $-1.20$ \\
\bottomrule
\end{tabular}
\end{table}

\subsection{Expert benchmark and its audit}
\label{sec:res-bench}

\begin{figure}[ht]
\centering
\includegraphics[width=\linewidth]{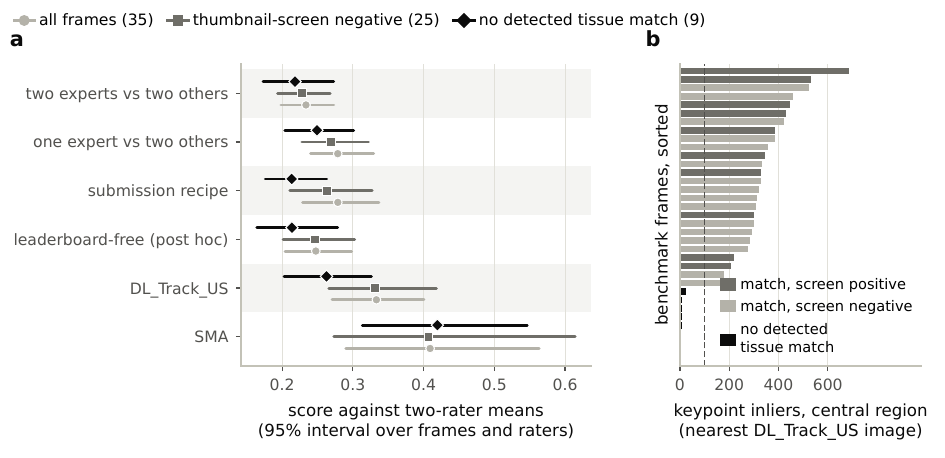}
\caption{Expert benchmark. (a)~Score against two-rater means on three frame sets, with 95\% intervals over frames and raters; the leaderboard-free reading was identified after the results were seen. (b)~Keypoint inliers between the central region of each benchmark frame and its nearest DL\_Track\_US training image; the dashed line separates tissue matches from frames with no detected tissue match, which also have none in the challenge training images. Screen positive and negative refer to the thumbnail screen.}
\label{fig:bench}
\end{figure}

Matching the central tissue region gave two separate groups (Figure~\ref{fig:bench}b; Supplementary Table~\ref{tab:nearcopy}): 26 frames had 164 to 688 inlier keypoints with one DL\_Track\_US training image, and 9 had 25 or fewer. Matched pairs showed the same tissue with different interface text. The thumbnail screen had flagged 10 of the 26, so 16 of the 25 thumbnail-screen-negative frames are tissue matches. The challenge training images split the benchmark the same way: the same 26 frames matched a challenge image with 71 to 688 inliers, and the other 9 reached at most 11 inliers. The segmentation models of this pipeline were trained on these images, so on the 26 frames DL\_Track\_US and the pipeline are both scored on tissue they have seen. Matched against every training image of both sets without the candidate preselection, the 9 frames reached at most 25 inliers with a DL\_Track\_US image and 11 with a challenge image as stored, and at most 9 mirrored in either set (Supplementary Section~\ref{app:nearcopy}). They have no detected tissue-level match in either training set, in either orientation, and all come from one screen layout, 660 pixels wide and 556 high at 7.7 pixels per millimetre.

On the 9 frames with no detected tissue match the recipe scored 0.213 (0.176 to 0.263), two experts against two others 0.218 and DL\_Track\_US 0.263 (Table~\ref{tab:bench}); both differences of the recipe include zero. With strict pairs, which need both raters of a reference pair, the two-expert reference fell to 0.196 while the recipe (0.209) and DL\_Track\_US (0.259) changed little, so the recipe's near parity with the experts on these frames rests partly on the single-rater fallback; the strict difference, $+0.013$ ($-0.048$ to 0.086), still includes zero (Supplementary Table~\ref{tab:strict}). In Table~\ref{tab:bench}, resampling frames alone gives about half the width of the interval for the recipe minus two experts on the 9 frames ($-0.036$ to 0.028 against $-0.058$ to 0.062); the draw of raters gives the rest. On the 25 thumbnail-screen-negative frames the recipe was 0.036 behind the two-expert reference ($-0.015$ to 0.094). The leaderboard-free reading scored 0.247 there, mainly because it drops the angle offset and the shrinkage. Nine frames support no claim of equivalence or non-inferiority. Per target (Table~\ref{tab:agree}), the pipeline's thickness was closer to the panel than that of DL\_Track\_US on both sets, and DL\_Track\_US was closer in length on the 25 frames. Whether the thickness bias of DL\_Track\_US comes from its segmentation or from its definition of thickness was not examined.

\begin{table}[ht]
\centering
\caption{Benchmark score against two-rater means, with 95\% intervals in brackets from resampling frames and 5 of 7 raters. Thumbnail-screen negative: frames not flagged by the thumbnail correlation, 16 of which are tissue matches. No detected match: frames with no detected tissue match in the DL\_Track\_US or the challenge training images. Sheet: readings given in the benchmark results sheet. Rater R5 read 17 frames, so some pair means rest on one rater; Supplementary Table~\ref{tab:strict} repeats the table with strict pairs. Differences are computed before rounding.}
\label{tab:bench}
\footnotesize
\setlength{\tabcolsep}{4.5pt}
\begin{tabular}{l l l l}
\toprule
& All frames & Thumbnail-screen & No detected \\
Reading & ($n = 35$) & negative ($n = 25$) & tissue match ($n = 9$) \\
\midrule
Submission recipe & 0.279 [0.229, 0.337] & 0.264 [0.211, 0.327] & 0.213 [0.176, 0.263] \\
Leaderboard-free (post hoc) & 0.248 [0.204, 0.298] & 0.247 [0.201, 0.302] & 0.214 [0.164, 0.279] \\
DL\_Track\_US (sheet) & 0.333 [0.271, 0.401] & 0.332 [0.266, 0.418] & 0.263 [0.203, 0.326] \\
SMA (sheet) & 0.409 [0.290, 0.563] & 0.407 [0.273, 0.614] & 0.419 [0.313, 0.547] \\
One expert vs two others & 0.279 [0.240, 0.329] & 0.270 [0.228, 0.322] & 0.249 [0.204, 0.301] \\
Two experts vs two others & 0.234 [0.198, 0.273] & 0.228 [0.194, 0.268] & 0.218 [0.173, 0.273] \\
\midrule
Recipe minus DL\_Track\_US & $-0.055$ [$-$0.139, 0.025] & $-0.068$ [$-$0.162, 0.021] & $-0.049$ [$-$0.125, 0.026] \\
Recipe minus two experts & $+0.045$ [$-$0.011, 0.107] & $+0.036$ [$-$0.015, 0.094] & $-0.005$ [$-$0.058, 0.062] \\
Leaderboard-free minus two experts & $+0.014$ [$-$0.032, 0.069] & $+0.019$ [$-$0.026, 0.075] & $-0.004$ [$-$0.063, 0.075] \\
 
\bottomrule
\end{tabular}
\end{table}

\begin{table}[ht]
\centering
\caption{Agreement with the panel mean per target: mean absolute error, mean difference (reading minus panel) and root-mean-square error, in degrees for PA and millimetres for FL and MT. The leaderboard-free reading is \col{pa\_deg} without offset, \col{fl\_lin3} and \col{mt\_mm}; its thickness is the recipe's, so the two MT rows are the same.}
\label{tab:agree}
\small
\begin{tabular}{l l r r r r r r}
\toprule
& & \multicolumn{3}{c}{\shortstack{Thumbnail-screen\\negative ($n = 25$)}} & \multicolumn{3}{c}{\shortstack{No detected\\tissue match ($n = 9$)}} \\
\cmidrule(lr){3-5}\cmidrule(lr){6-8}
Target & Reading & MAE & Bias & RMSE & MAE & Bias & RMSE \\
\midrule
\multirow{3}{*}{PA (deg)} & recipe & 1.27 & $+0.91$ & 1.57 & 1.43 & $+1.43$ & 1.73 \\
 & leaderboard-free & 1.01 & $-0.29$ & 1.31 & 0.75 & $+0.23$ & 1.00 \\
 & DL\_Track\_US & 1.44 & $-1.30$ & 1.72 & 1.56 & $-1.20$ & 1.93 \\
\midrule
\multirow{3}{*}{FL (mm)} & recipe & 4.90 & $+0.99$ & 6.51 & 2.83 & $+2.47$ & 3.39 \\
 & leaderboard-free & 4.32 & $+2.62$ & 5.59 & 3.92 & $+2.37$ & 5.00 \\
 & DL\_Track\_US & 3.74 & $+0.36$ & 4.98 & 3.23 & $-0.09$ & 4.42 \\
\midrule
\multirow{3}{*}{MT (mm)} & recipe & 0.23 & $+0.13$ & 0.29 & 0.15 & $+0.14$ & 0.19 \\
 & leaderboard-free & 0.23 & $+0.13$ & 0.29 & 0.15 & $+0.14$ & 0.19 \\
 & DL\_Track\_US & 1.03 & $+0.87$ & 1.67 & 0.45 & $+0.37$ & 0.77 \\
 
\bottomrule
\end{tabular}
\end{table}

\subsection{Quality flags and contribution to the blend}
\label{sec:res-fail}
\label{sec:res-own}

The flags caught 3 of the 10 benchmark frames with a large error and raised 5 false alarms (sensitivity 0.30, specificity 0.80); they can only serve as auxiliary signs and not to accept or reject a reading automatically. The three public notebooks agreed with one another at correlations of 0.978 to 1.000 across the three targets, and the pipeline's predictions correlated with them at 0.80 to 0.94 (Supplementary Figure~\ref{fig:sources}). Averaging the pipeline's angle with notebook M gained 0.01298 public and 0.01354 private, and averaging the lengths 0.01654 and 0.02625, while the pipeline's thickness was worse than M's on both splits.

\subsection{Leaderboard decisions on the private split}
\label{sec:res-replication}

\begin{figure}[ht]
\centering
\includegraphics[width=\linewidth]{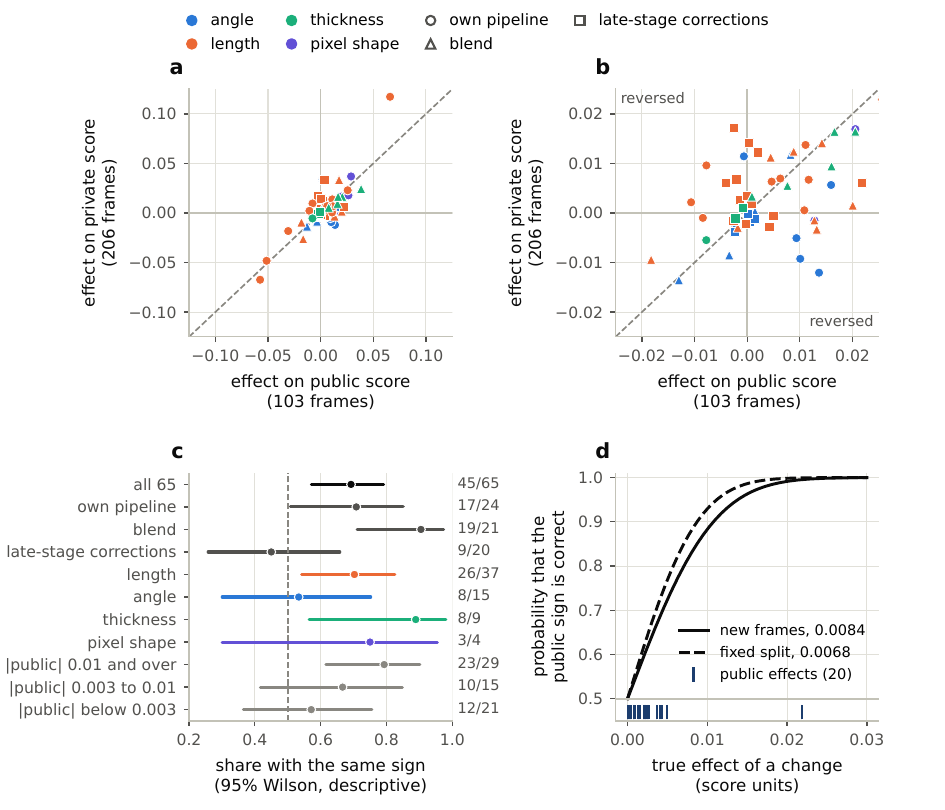}
\caption{Replication of 65 single-change contrasts; negative values are improvements. (a)~Public against private effect, coloured by target, marked by stage. (b)~The same near zero. (c)~Share of contrasts with the same sign, with Wilson intervals that ignore the dependence between contrasts and are descriptive only. (d)~Probability that the public sign is right, given the true effect, when the RMS-based dispersion scale of the late-stage corrections is read as a standard error; an illustration under that assumption, not a calibrated model. Ticks mark the public effects of the 20 late-stage corrections.}
\label{fig:replication}
\end{figure}

Additivity held on the private split: \entry{n\_comb}, four changes of \entry{h\_best} taken together, was predicted from the scores of those four submissions and of \entry{h\_best} at 0.24152 public and 0.27993 private, the scores it received (Supplementary Section~\ref{app:replication}). Across the 65 contrasts of this entry (Figure~\ref{fig:replication}, Table~\ref{tab:repset}) public and private effects correlated at $r = 0.81$ (cluster bootstrap 0.39 to 0.93) and agreed in sign in 45 cases (69\%; Wilson interval 57 to 79\%, cluster bootstrap 58 to 86\%). Agreement depended on the kind and size of a change. Pipeline and blend changes kept their sign in 17 of 24 and 19 of 21 cases, the 20 late-stage corrections, global and layout-specific, in 9 ($r = 0.11$). Thickness and length changes replicated ($r = 0.97$ and 0.87), angle changes did not ($r = 0.12$). Changes with a public effect of 0.01 or more kept their sign in 23 of 29 cases, and those below 0.003 in 12 of 21.

\begin{table}[ht]
\centering
\caption{Public against private effects of the 65 contrasts by subset: correlations, contrasts with the same sign (Wilson interval, descriptive; contrasts sharing a base are not independent), mean and root-mean-square difference, and the RMS-based dispersion scale of a public effect, Eq.~\eqref{eq:se}, for new frames ($D_{\mathrm{new}}$) and for the fixed split ($D_{\mathrm{fixed}}$); these are not conventional standard errors.}
\label{tab:repset}
\footnotesize
\setlength{\tabcolsep}{4.2pt}
\begin{tabular}{l r r r l r r r r}
\toprule
Subset & $n$ & $r$ & Spearman & Same sign & Mean diff. & RMS diff. & $D_{\mathrm{new}}$ & $D_{\mathrm{fixed}}$ \\
\midrule
\multicolumn{9}{l}{\textit{All changes}} \\
all & 65 & 0.81 & 0.58 & 45 (57--79\%) & $-0.0008$ & 0.0125 & 0.0102 & 0.0083 \\
\multicolumn{9}{l}{\textit{By stage}} \\
own pipeline & 24 & 0.87 & 0.62 & 17 (51--85\%) & $-0.0016$ & 0.0164 & 0.0134 & 0.0109 \\
blend & 21 & 0.80 & 0.81 & 19 (71--97\%) & $+0.0025$ & 0.0088 & 0.0072 & 0.0059 \\
late-stage corrections & 20 & 0.11 & 0.04 & 9 (26--66\%) & $-0.0034$ & 0.0103 & 0.0084 & 0.0068 \\
\multicolumn{9}{l}{\textit{By target}} \\
pixel shape & 4 & 0.92 & 1.00 & 3 (30--95\%) & $+0.0046$ & 0.0094 & 0.0077 & 0.0063 \\
angle & 15 & 0.12 & 0.20 & 8 (30--75\%) & $+0.0024$ & 0.0130 & 0.0106 & 0.0087 \\
length & 37 & 0.87 & 0.53 & 26 (54--83\%) & $-0.0035$ & 0.0137 & 0.0112 & 0.0092 \\
thickness & 9 & 0.97 & 1.00 & 8 (56--98\%) & $+0.0022$ & 0.0056 & 0.0046 & 0.0037 \\
\multicolumn{9}{l}{\textit{By size of the public effect}} \\
0.01 and over & 29 & 0.88 & 0.87 & 23 (62--90\%) & $+0.0033$ & 0.0149 & 0.0122 & 0.0100 \\
0.003 to 0.01 & 15 & 0.27 & 0.28 & 10 (42--85\%) & $-0.0032$ & 0.0107 & 0.0088 & 0.0072 \\
below 0.003 & 21 & 0.35 & 0.34 & 12 (37--76\%) & $-0.0047$ & 0.0097 & 0.0079 & 0.0064 \\
 
\bottomrule
\end{tabular}
\end{table}

For the late-stage corrections the dispersion scale of a public effect was 0.0084 for new frames and 0.0068 for the fixed split. The global length scale of 1.045 alone gave 41\% of the sum of squares; without it the scales were 0.0066 and 0.0054. Read as standard errors, these values mean that a public effect needed about 0.011 to 0.017 to differ from zero at the 5\% level, and that for the public effects of 0.002 to 0.004 on which most corrections were chosen the public sign is right with a probability of about 0.6 to 0.7; Figure~\ref{fig:replication}d is an illustration under this reading, not a calibrated model. The corrections in \entry{n\_pa\_tm\_up}, the per-layout terms and the thickness offset, improved the public score by 0.00921 and cost 0.00025 on the private split. They were the best of about 20 tried, so their public gain carries a selection bias, the winner's curse; without true effects their expected private gain is zero. A global length scale of 1.015, the public optimum, gained 0.00406 public and cost 0.00613 private, so \entry{p\_safe}, built as a guard against overfitting, scored worse on the private split than the per-layout entry. One large decision also reversed: the old fascicle model was kept over newer five-fold models on public losses of 0.01364 and 0.00935, and on the private split the newer angles gained 0.01205 and 0.00508.

\section{Discussion}
\label{sec:discussion}

\subsection{Frame geometry can alter measurements substantially}
\label{sec:disc-geometry}

The largest single improvement of the entry came from the frame, before any model changed. Equation~\eqref{eq:aspect} explains the size: at 15\textdegree, a Lumify frame read as square gains 2.8\textdegree{} of angle, nearly twice the spread between benchmark raters, and an oblique length shrinks by 15\%, while thickness does not change. A pipeline checked on thickness first can miss the problem. The organiser stated that no image has different horizontal and vertical resolution \citep{umud689019}. This and the private-split gain can both hold if the raters measured the frames in their native geometry and the stored frames were later fitted to a canvas of another shape, but the data cannot test this. What they show is narrower: for these inputs a layout-derived correction improved scores on frames that played no part in choosing it, for the Lumify layout and for the Siemens and Lumify frames together; for the Telemed 800 layout it did not, in a test confounded with the unshrunk closed form. The Monte Carlo analysis points to the vertical scale as the reading that needs most care, because a scale error of 2\% already exceeds the raters' spread on thickness, the target with the largest weight per millimetre.

\subsection{Fascicle extrapolation is a major source of length uncertainty}

On flat fascicles the extension rule is part of the label: the two conventions differed by more than the tolerance on about half of the test frames. The least-extrapolated reading agreed best with the benchmark raters and with the private split, in the same direction as reports that linear extension overestimates length against extended field-of-view scans \citep{pimenta2018}. No reading here is validated against a full-length reference. The shrinkage of Eq.~\eqref{eq:trust} lowered the mean absolute error, but it pulled every length towards a value above all panel lengths; under an absolute-error score and a skewed $\FL_c$, a pull of this kind can lower the score while biasing single frames. The weight $w$ is a heuristic chosen for the challenge score and is not a calibrated uncertainty. A measurement tool should estimate uncertainty from independently labelled data, with prediction intervals whose coverage has been checked; until then, $w$ can only show how much a length leans on the geometric closed form.

The pipeline's thickness was the most accurate reading on the benchmark and worse than notebook M's on both leaderboard splits. The thickness direction does not explain the gap: $\MT_\perp$ differed from $\MT_v$ on the test frames by a median of 0.03~mm, while the pipeline and notebook M differed by 0.48~mm on average and both read vertical gaps. Nor does the border convention: on the pipeline's own masks, the centre-line reading of notebook M (\col{mt\_ch955}) differed from the inner-edge gap by 0.06~mm on average. The gap therefore lies in the segmentation weights and in the scale each notebook reads from the screen, and the benchmark, with its given scale and without the rectus femoris or participants with cerebral palsy, cannot tell which of the two is right on the test frames. On its own the pipeline trailed notebook M; M averages two segmentation ensembles while the pipeline's fascicle map came from one model trained in a single fold, though neither difference was tested alone. The pipeline helped the blend because its errors differed from those of the three public files, which share weights and code.

\subsection{Public-leaderboard optimisation can fail on the private split}

Within this entry, large decisions held on the private split and small ones did not. If the RMS-based dispersion scale of 0.005 to 0.008 is read as a standard error, a public effect of 0.003 points the right way with a probability of about 0.6 to 0.7, and the corrections kept for their public gain are biased upward by selection. The additive score made effects measurable to five decimals, but that precision concerns 103 frames. This is adaptive overfitting \citep{blum2015,dwork2015} measured on one entry, and a guard built from the same split, \entry{p\_safe}, did not protect against it. The threshold of about 0.01 above which effects tended to replicate holds for this entry and score, not as a general rule.

\subsection{Benchmark use and recommendations}

The audit establishes tissue-level overlap between 26 benchmark frames and the training images of both DL\_Track\_US and the challenge, so on those frames DL\_Track\_US and the present pipeline are both scored on tissue their models have seen; only the raters are unaffected. The nine frames with no detected match in either set are the part of the benchmark free of detected overlap, but they come from one screen layout and, without participant identifiers, may still share participants with the training data, so they are not an independently verified validation set. Global similarity measures missed most matches because interface text changed between versions of a frame, and the benchmark stores frames in both left-right orientations; matching the central tissue region separated the groups cleanly. Nine frames are also too few to rank the methods. These results suggest a few practical steps: store the pixel spacing along both axes with every frame; publish the extension rule and the thickness direction with the labels; audit benchmarks for tissue-level overlap with public training sets, in both orientations, before release; and report a noise estimate with a leaderboard and split it by recording (Supplementary Table~\ref{tab:recs}).

\section{Limitations}
\label{sec:limitations}

The pixel aspect values were inferred from screen elements and not checked against metadata or a phantom, two of them were assumed, and they apply to the six layouts of this challenge, not to other scanner models or export settings. The sensitivity analysis draws aspect and scale errors independently; correlated errors from a common resizing or export step were not modelled. Benchmark evidence free of detected training overlap rests on nine frames of one screen layout, without the rectus femoris or participants with cerebral palsy and with the scale given. These frames were compared with both training sets in both orientations, but without participant identifiers they may still share participants with the training data, and on them the human reference depends on how missing rater entries are handled. The segmentation models were validated only on folds that group near-identical frames, not participants, without IoU or boundary metrics, and the fascicle checkpoint is not tied to its log by a hash. No length reading is validated against a full-length reference, and the raters' chosen fascicles are not recorded.

The replication rests on one entry's submissions: 65 contrasts from 13 bases, not a random sample of changes, with an unknown frame-to-split assignment and no verified participant or recording identifiers; the recordings used here were inferred from the images. The private split is itself a sample, so it separates large from small effects without confirming any single small one, and the dispersion scales of Eq.~\eqref{eq:se} mix noise with differences between true effects. The recording median and the layout fallback for four failed frames were not ablated, although the saved readings bound the effect of the median, the public notebooks were not run on the benchmark, and residual correlations between the blended files need test labels that are not released.

Reproduction has one gap. The tables and figures rebuild from the derived files, the blended files and both selected entries from the pipeline file and the notebook outputs, and the pipeline's test-set readings from the raw images with the published code and weights, byte for byte; but \entry{b\_c93} is a historical file assembled from two runs and is not rebuilt by one, so it is published as submitted. The output of notebook T is tied to its version by run time, not by a second hash check, and the version of \texttt{segmentation-models-pytorch} is inferred, not recorded. The data support no claim about other muscles, scanners, contracted muscle or clinical use.

\section{Conclusion}
\label{sec:conclusion}

In the UMUD Challenge, frame geometry and the extension rule for fascicles moved automated readings of muscle architecture as much as the choice of model. A layout-derived geometric correction improved private scores for some layouts and not for others. The two extension conventions differed by more than the tolerance on flat fascicles, which shows how uncertain single-frame length is. Corrections with small public effects often failed on the private split. Most frames of the public expert benchmark share their tissue with the training images of DL\_Track\_US and of the challenge, which limits its use as independent evidence. These findings come from one challenge entry and a small benchmark; their generality and anatomical validity need testing on independently acquired data with physical calibration references.

\section*{Data availability}

The challenge data are available to registered participants on the competition page under the competition rules \citep{umudkaggle}. The expert benchmark is available on OSF under CC BY 4.0 \citep{umudosf}. The derived data needed to rebuild every table and figure of this paper are published as the Kaggle dataset \texttt{umud kooklan derived data} \citep{kooklan2026data} and as ancillary files of this preprint: the pipeline's per-frame test readings without labels, the decisive submitted files, the list of scored submissions with their public and private scores, the benchmark readings and the overlap audits. The only images among them are benchmark frames and overlays drawn on benchmark frames; no test frame is included. The files made from the challenge data are shared for non-commercial research under CC BY-NC-SA 4.0, the licence of the challenge data. Test frames are not shown in this paper; all frames in the figures are benchmark frames.

\section*{Code availability}

The \texttt{umud} package (version 0.18.1, GPL-3.0-or-later) and the analysis scripts are published as the Kaggle dataset \texttt{umud kooklan code} \citep{kooklan2026code} and as ancillary files of this preprint, with the project page at \url{https://soheilkooklan.github.io}. The six trained checkpoints are published as \texttt{umud kooklan weights} \citep{kooklan2026weights}. The ancillary files hold the final version of the analysis scripts, the source of version 0.12.0 and the outputs of the Kaggle audit and reproduction runs. The training, prediction, benchmark, audit and reproduction notebooks are public on Kaggle; Supplementary Table~\ref{tab:kaggle} lists each with its version, and Supplementary Section~\ref{app:repro} gives the library versions and the SHA-256 hashes of the decisive files. The public notebooks used in the blend belong to their authors and are available on Kaggle \citep{musawwar2026,trying2026,bafna2026}; their output files are identified by hash and are not redistributed.

Three levels of reproduction were carried out. Statistical reproduction rebuilds every table and figure from the derived files on a CPU in minutes. Submission reproduction rebuilds the blended files and both selected entries, byte for byte, from the pipeline file \texttt{b c93.csv} and the three notebook outputs; \texttt{b c93.csv} itself is a historical file assembled from two runs, published as submitted. Prediction reproduction reran the pipeline's prediction step on the challenge test images with the published code (0.18.1, whose prediction files are unchanged from the 0.18.0 of the original run) and weights on Kaggle, and gave per-frame reading files identical, by SHA-256, to those of the original run.

\section*{Author contributions}

S.K. is the sole author and took part in the challenge as an individual participant. He designed the pipeline, wrote its code, trained the segmentation models, made all Kaggle submissions, designed and ran the analyses and wrote the manuscript, with the assistance described in the Acknowledgements.

\section*{Funding}

This work received no specific funding.

\section*{Ethics}

This work used only de-identified images released by the challenge organiser and the public UMUD benchmark; no new data were collected from human participants. According to the competition overview, the training images were sampled from earlier publications and are covered by their ethical approvals, and the test images were anonymised \citep{umudkaggle}. The benchmark frames are published under CC BY 4.0 by their authors \citep{umudosf,ritsche2026}.

\section*{Competing interests}

The author declares no competing interests. He took part in the challenge for research and for training in ultrasound image analysis and received no payment or prize for it.

\section*{Acknowledgements}

The blend uses the outputs of the public notebooks of Mawra Musawwar, Rishabh Bafna and prvsiyan. The texture angle and the centre-line thickness are re-implemented from Mawra Musawwar's notebook, the segmentation recipe follows watehell's notebook, and the layout rules and pixel aspect values follow the public analysis of AmbrosM and the notebook of watehell \citep{ambrosm2026,watehell2026}; a public notebook by honghanhhh used during the competition is no longer listed on Kaggle. Paul Ritsche organised the challenge, answered participants' questions and published the expert benchmark with his co-authors.

AI assistants were used in this work: Claude (Anthropic) debug code, edit the text, and ChatGPT (OpenAI) and Grok (xAI) to review the paper and the code. The author checked the code, the reported numbers and the references, and takes full responsibility for the content. No AI tool is an author of this paper.

{\small
\singlespacing
\bibliographystyle{unsrtnat}
\bibliography{references}
}

\clearpage
\singlespacing
\setcounter{section}{0}
\setcounter{table}{0}
\setcounter{figure}{0}
\renewcommand{\thesection}{S\arabic{section}}
\renewcommand{\thetable}{S\arabic{table}}
\renewcommand{\thefigure}{S\arabic{figure}}
\renewcommand{\theHsection}{S\arabic{section}}
\renewcommand{\theHtable}{S\arabic{table}}
\renewcommand{\theHfigure}{S\arabic{figure}}

\begin{center}
{\Large\bfseries Supplementary material\par}
\vspace{0.4em}
{\normalsize \papertitle\par}
\end{center}

\section{Segmentation and measurement}
\label{app:seg}

\begin{table}[H]
\centering
\begin{threeparttable}
\caption{Training settings of the aponeurosis models, from the training log (notebook \texttt{umud v2 train apo}, code 0.14.0) and \texttt{scripts/train.py}, whose training files are identical in 0.14.0 and 0.18.1.}
\label{tab:train}
\small
\begin{tabular}{l >{\raggedright\arraybackslash}p{11.6cm}}
\toprule
Setting & Value \\
\midrule
Architecture & U-Net, ResNet-34 encoder, ImageNet initialisation, one input channel, one output \\
Input & greyscale, $384 \times 576$ pixels, per-image standardisation \\
Data & 1048 frames with aponeurosis masks, no duplicates dropped; 866 hash groups, the largest of 14 frames \\
Folds & 5, grouped by difference hash (20 bits; 48 bits for consecutive files), 838 or 839 training and 209 or 210 held-out frames; not grouped by participant \\
Optimiser & AdamW, learning rate $3 \times 10^{-4}$, weight decay $10^{-4}$, one-cycle schedule \\
Schedule & 30 epochs, batch size 8, mixed precision \\
Loss & $0.5\,$BCE $+ 0.5\,$Dice; BCE positive weight $(1-c)/c$ from the median mask coverage $c$, capped at 200 (31.0 here); masks not dilated \\
Augmentation & independent random crop of each axis to 60--100\%; horizontal flip ($p = 0.5$); rotation within $\pm 7^\circ$ ($p = 0.5$); gamma 0.7--1.5 ($p = 0.5$); contrast 0.75--1.25 and brightness $\pm 0.1$ ($p = 0.7$); multiplicative noise ($p = 0.3$); Gaussian blur ($p = 0.1$) \\
Seed & 1337 for fold assignment; 1337 plus the fold number for initialisation and data order \\
Threshold & per fold, matched to annotated mask thickness on held-out frames (0.95--0.975); mean 0.955 used \\
Test time & mean of five folds, with and without horizontal flip \\
\bottomrule
\end{tabular}
\end{threeparttable}
\end{table}

The fascicle model is a single-fold U-Net of the same design at $512 \times 512$ pixels, trained with version 0.12.0 of the code and applied at a threshold of 0.9. Its training log records that 814 duplicate frames were removed from the 2761 fascicle frames, leaving 1947 image--mask pairs; fold 0 of a blocked split used 1558 of them for training and 389 for validation. The blocked split sorts the pairs by file number, cuts them into 40 contiguous blocks and holds out a fifth of the blocks, so that neighbouring files, which tend to come from one recording, stay on one side; it does not group by participant. Training ran for 35 epochs with seed 1337, masks dilated by 5 pixels and a positive weight of 61.2 in the binary cross-entropy, and the best validation Dice was 0.488 at a threshold of 0.9. The checkpoint used in the entry is the output of the notebook \texttt{train-fasc} (version 1), published with the other weights; the log summarised here was kept separately, and no hash ties it to that checkpoint.

\begin{table}[H]
\centering
\caption{Validation records of the segmentation models. Dice is pooled over the held-out pixels at the best of a grid of thresholds, and the best epoch was chosen on the same frames, so the values are optimistic; IoU and boundary metrics were not computed. Mask-derived angles come from the annotated training masks, not from raters. Orientation network: a U-Net trained to predict the fascicle direction field from the annotations; ridge filter: a projection search over angles on a ridge-enhanced image; constant: the median training angle for every frame.}
\label{tab:segval}
\small
\begin{tabular}{>{\raggedright\arraybackslash}p{4.0cm} >{\raggedright\arraybackslash}p{4.3cm} >{\raggedright\arraybackslash}p{6.2cm}}
\toprule
Model & Validation set & Result \\
\midrule
Aponeurosis, five folds (used) & 209 or 210 held-out frames per fold & best Dice 0.8125, 0.7924, 0.7978, 0.8119 and 0.7979 (mean 0.80) at epochs 26, 29, 29, 30 and 25; thresholds 0.95--0.975 matched to mask thickness \\
Aponeurosis, five folds (used) & benchmark frames & vertical gap 0.23~mm from the panel mean on the 25 thumbnail-screen-negative frames, 0.15~mm on the 9 frames with no detected tissue match \\
Fascicle, single fold, code 0.12.0 (used) & fold 0, 389 held-out pairs & best Dice 0.488 at threshold 0.9, from a log kept separately and not tied to the checkpoint by a hash \\
Fascicle, single fold, code 0.10.0, with the fragment angle & 179 held-out frames, angle of the annotated masks & mean absolute error 1.00\textdegree; orientation network 1.44\textdegree, ridge filter 3.43\textdegree, constant 5.16\textdegree \\
\bottomrule
\end{tabular}
\end{table}

\begin{table}[H]
\centering
\begin{threeparttable}
\caption{Measurement columns written by the pipeline for every frame. Origin: own, developed for this work; adapted, re-implemented from the public notebook cited.}
\label{tab:columns}
\small
\begin{tabular}{l >{\raggedright\arraybackslash}p{9.1cm} l}
\toprule
Column & Definition & Origin \\
\midrule
\col{pa\_deg} & length-weighted median fragment angle against the local deep tangent & own \\
\col{pa\_deep} & linear fit of fragment angle against relative depth, weighted by (length)$^{1/2}$, read at depth 0.9 & own \\
\col{pa\_texture} & direction of maximum projection variance of the high-pass filtered belly & adapted \citep{musawwar2026} \\
\col{pa\_ins3} & mean angle of the three longest fragments in the deep half of the belly, against the straight deep line & own \\
\col{pa\_lmed} & median fragment angle against the straight deep line & own \\
\col{fl\_mm} (recipe) & closed form $\MT/\sin\PA$ shrunk by Eq.~\eqref{eq:trust} & own \\
\col{fl\_extend} & fragments extended along their own line to the curved inner edges, length-weighted median & own \\
\col{fl\_centre} & fragments extended to the band centre lines, length-weighted median & adapted \citep{musawwar2026,watehell2026} \\
\col{fl\_lin3} & mean full length of the three least-extrapolated fragments, straight aponeuroses & own \\
\col{fl\_lmed} & median full length over all fragments, straight aponeuroses & own \\
\col{mt\_mm} & vertical inner gap at 20, 50, 80\% of the shared width, threshold 0.955 & own \\
\col{mt\_a900}, \col{mt\_a975} & the same at thresholds 0.9 and 0.975 & own \\
\col{mt\_ch955}, \col{mt\_ch500} & centre-line gap less half the band heights, thresholds 0.955 and 0.5 & adapted \citep{musawwar2026} \\
\col{mt\_perp} & shortest distance from the superficial curve to the deep curve at the same columns, averaged ($\MT_\perp$; analysis only) & own \\
\bottomrule
\end{tabular}
\end{threeparttable}
\end{table}

\section{Sensitivity of the layout-derived correction}
\label{app:mc}

\begin{figure}[H]
\centering
\includegraphics[width=\linewidth]{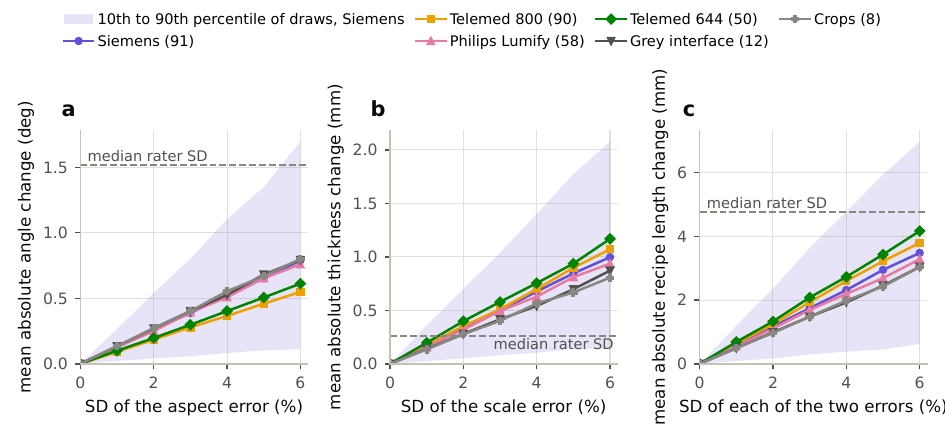}
\caption{Sensitivity of the pipeline's test readings to a calibration error systematic within a layout (2000 draws per point, all 309 frames, no labels). (a)~Angle against an error in $a$; (b)~thickness against an error in $v$; (c)~recipe length with both errors. Lines are means over draws; the band is the 10th to 90th percentile for the Siemens layout; dashed lines mark the median standard deviation between benchmark raters.}
\label{fig:calmc}
\end{figure}

A calibration error is systematic within a layout. For each layout, 2000 pairs of relative errors $(e_a, e_v)$ were drawn from normal distributions with standard deviations of 0 to 6\% and applied to all its frames, the four failed frames with their layout median included: the frame angle $\theta$, the reading \col{pa\_deg} before the $+1.2^\circ$ offset, becomes $\arctan\{(1+e_a)\tan\theta\}$ with the deep aponeurosis taken as horizontal, thickness is divided by $1+e_v$, and the recipe length is recomputed from the perturbed angle plus the offset. \col{fl\_lin3} is not measured again on resampled masks; it is multiplied by the length factor $\{\sin^2\theta/(1+e_v)^2 + \cos^2\theta/[(1+e_v)(1+e_a)]^2\}^{1/2}$ at the frame angle. An aspect error of 2\% moved the angle by 0.19 to 0.27\textdegree{} and 5\% by 0.46 to 0.68\textdegree; a scale error of 2\% moved thickness by 0.28 to 0.40~mm and 5\% by 0.67 to 0.94~mm. With both errors at 5\%, the recipe length moved by 2.4 to 3.4~mm, the unshrunk closed form $\FL_{c,v}$ by 2.6 to 5.3~mm and \col{fl\_lin3} by 3.4 to 5.5~mm. The shrinkage damps the sensitivity mainly through the factor $w < 1$ on the closed form, since the share $1-w$ anchored to $F_0$ does not move with the calibration. Differentiating Eq.~\eqref{eq:trust} in the angle gives $w\,\partial\FL_{c,v}/\partial\PA + (\FL_{c,v} - F_0)\,\partial w/\partial\PA$, where the first term is negative and $\partial w/\partial\PA$ is positive. The second term therefore partly cancels the first where $\FL_{c,v} > F_0$, as on the flat Telemed layouts (median $\FL_{c,v}$ of 95 and 105~mm, $w$ about 0.65 and 0.68), and adds to it where $\FL_{c,v} < F_0$, as on the steeper layouts (median $\FL_{c,v}$ 50 to 68~mm, $w$ about 0.8). The recipe line in Figure~\ref{fig:calmc}c therefore understates the sensitivity of an unshrunk length. Reading the Telemed 800 frames as square would raise their angles by 1.16\textdegree{} and shorten the recipe length by 5.0~mm, the unshrunk closed form without the offset by 9.6~mm and \col{fl\_lin3} by 8.6~mm.

\begin{table}[H]
\centering
\begin{threeparttable}
\caption{Median pipeline readings per layout on the 309 test frames, after the recording median and before offsets and blending; no labels are involved. Frames: all frames of the layout; Failed: frames without a pair of aponeuroses, which carry their layout median and leave the medians unchanged at the precision shown (Figure~\ref{fig:conventions} excludes them and shows 305 frames). $|\beta|$ and ratio are taken over the 295 frames whose saved overlays allowed both aponeuroses to be refitted. Gap: \col{fl\_centre} minus \col{fl\_lin3}. $|\beta|$: median inclination of the deep aponeurosis; ratio: median perpendicular over vertical thickness.}
\label{tab:testlayouts}
\small
\begin{tabular}{l r r r r r r r r r}
\toprule
Layout & Frames & \col{pa\_deg} & \col{fl\_lin3} & \col{fl\_centre} & Gap & \col{mt\_mm} & $|\beta|$ & Ratio & Failed \\
& & (deg) & (mm) & (mm) & (mm) & (mm) & (deg) & & \\
\midrule
Siemens & 91 & 17.8 & 73.2 & 81.2 & 8.1 & 20.7 & 3.0 & 0.998 & 0 \\
Telemed 800 & 90 & 12.2 & 95.7 & 114.8 & 19.1 & 22.4 & 2.4 & 0.999 & 0 \\
Philips Lumify & 58 & 16.9 & 67.4 & 73.2 & 5.8 & 19.6 & 3.0 & 0.998 & 3 \\
Telemed 644 & 50 & 13.1 & 90.7 & 108.6 & 17.9 & 25.9 & 2.5 & 0.999 & 0 \\
Grey interface & 12 & 19.3 & 59.4 & 64.1 & 4.7 & 17.2 & 7.9 & 0.991 & 1 \\
Crops & 8 & 18.2 & 62.9 & 67.9 & 5.0 & 17.6 & 8.1 & 0.990 & 0 \\
\bottomrule
\end{tabular}
\end{threeparttable}
\end{table}

\section{Length, thickness and example frames}
\label{app:length}

\begin{figure}[H]
\centering
\includegraphics[width=\linewidth]{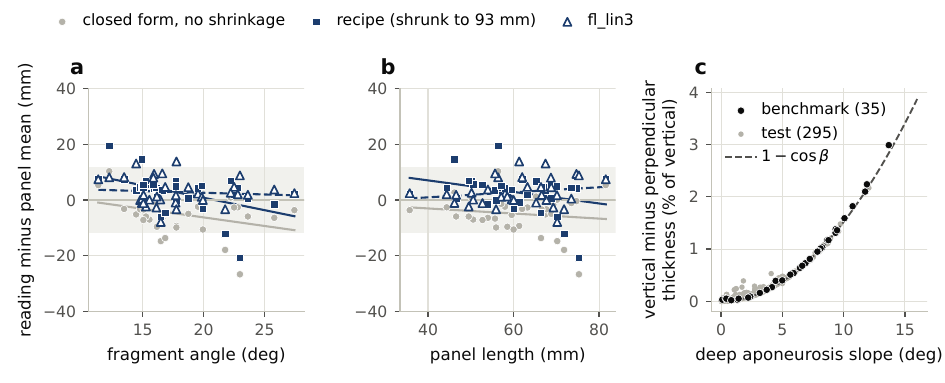}
\caption{Length errors on the 35 benchmark frames and the thickness convention. (a,b)~Reading minus panel mean against the fragment angle and the panel length, with least-squares lines; the band is $\pm 12$~mm. (c)~Relative difference between vertical and perpendicular thickness against the slope of the deep aponeurosis on benchmark frames and on the 295 refitted test frames, with the curve $1-\cos\beta$.}
\label{fig:lenfail}
\end{figure}

\begin{figure}[H]
\centering
\includegraphics[width=\linewidth]{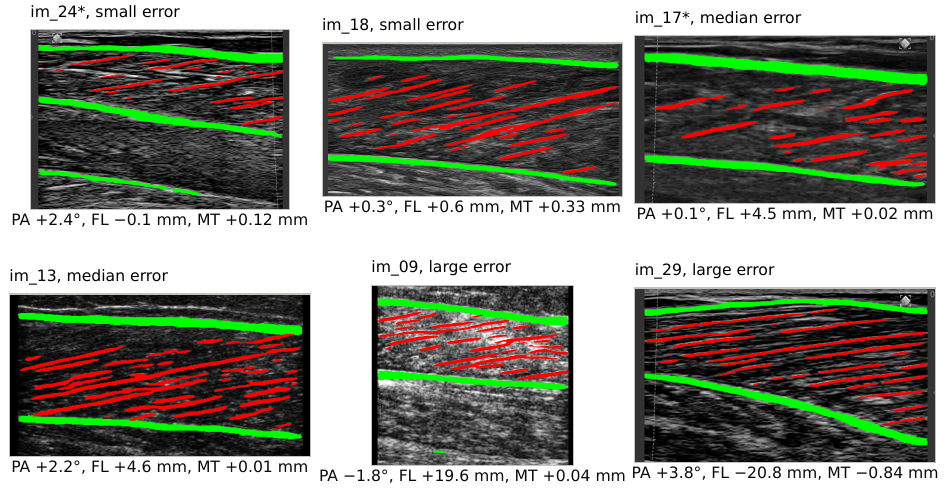}
\caption{Benchmark frames with small, median and large length errors of the recipe, with the aponeurosis (green) and fascicle (red) masks of the pipeline (OSF, CC BY 4.0). Errors against the panel mean are printed on each frame in degrees and millimetres; an asterisk marks frames with no detected tissue match.}
\label{fig:gallery}
\end{figure}

\begin{figure}[H]
\centering
\includegraphics[width=\linewidth]{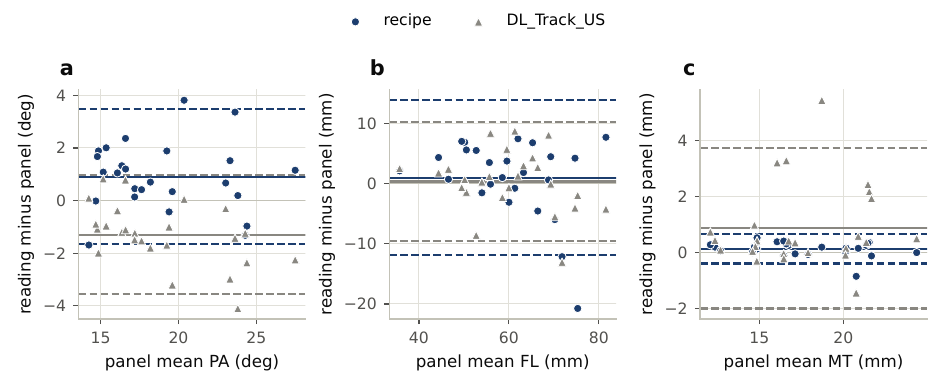}
\caption{Bland--Altman plots on the 25 thumbnail-screen-negative benchmark frames, for the recipe and DL\_Track\_US. Solid lines: mean difference; dashed lines: 95\% limits of agreement. Of these frames, 16 share their tissue with training images of DL\_Track\_US and the challenge. In (c) one DL\_Track\_US point lies at $+4.9$~mm, at the upper edge of the plot.}
\label{fig:ba}
\end{figure}

\section{Overlap audit of the benchmark}
\label{app:nearcopy}

Table~\ref{tab:nearcopy} gives, for each benchmark frame, the closest of the 3553 DL\_Track\_US training images under each measure and the central inliers of its best challenge training image. The dHash distance counts differing bits out of 256. The central-region matches leave no frame between 25 and 164 inliers in the DL\_Track\_US set and none between 11 and 71 in the challenge set, so the grouping does not depend on the exact limit; both sets give the same 26 matches.

\paragraph{Challenge training images} The comparison ran on Kaggle in the notebook \texttt{umud audit challenge train} (version 2). It read every image in the two training folders of the challenge, which hold the 2761 fascicle and 1048 aponeurosis frames, resized each to 512 pixels wide as for DL\_Track\_US, cut the same central half of the benchmark frame and used the same ORB, ratio-test and RANSAC settings and the same 20 plus 20 candidates; images present in both folders were not removed, since a repeated image cannot change a maximum count. The notebook saves each benchmark frame beside its best match.

\paragraph{Orientation and exhaustive search} Some benchmark frames are stored mirrored left to right, so their scanner text reads backwards; of the nine frames without a match, im\_23, im\_24 and im\_31 show their text the right way round and the other six reversed. None of the four measures is mirror invariant, and the candidate preselection could in principle miss a match, so the nine frames were matched again, as stored and mirrored, against every image of both sets. Against the DL\_Track\_US set they reached 5 to 25 inliers as stored and 5 to 9 mirrored (\texttt{out/near\_copy\_flip.csv}); against the challenge set, 5 to 11 as stored and 5 to 9 mirrored (\texttt{out/audit\_challenge\_flip.csv}, written by the notebook). The 26/9 grouping did not change. Table~\ref{tab:nearcopy} shows the counts with preselection, so im\_17 has 0 central inliers there and 5 in the exhaustive search.

\paragraph{Full frame against central region} Six of the nine frames have 99 to 196 full-frame inliers against 25 or fewer in the central region. Their full-frame matches come from the interface, rulers and borders that frames of one screen layout share; the central crop leaves only tissue, where these frames have no counterpart.

\paragraph{Matched pairs} For the 26 matched frames the RANSAC homography between the central half of the benchmark frame and its DL\_Track\_US match was fitted again (\texttt{out/near\_copy\_homography.csv}). Both images are resized with one factor for both axes, so the ratio of lateral to vertical scale measures any stretch between the two versions: its median was 1.003 (0.979 to 1.030), the scales lay between 0.97 and 1.02, and no pair was mirrored. The matched versions differ in their interface and crop, not measurably in geometry.

\paragraph{Raters and resampling} The results sheet holds seven rater columns. Rater R5 read only 17 of the 35 frames; R5 stays in all main analyses, and a sensitivity analysis without R5, added at a reviewer's request after the results were known, changed the recipe score on the thumbnail-screen-negative frames from 0.264 to 0.258 and the two-expert reference from 0.228 to 0.215. The slip rule set to missing any entry more than 50\% from the median of the other raters for that frame and target. It removed two entries of rater R7, both on thumbnail-screen-positive frames: the thickness of im\_19 (80.1~mm against a median of 20.0~mm for the other raters) and the length of im\_26 (33.9~mm against 70.0~mm). Keeping them raised the scores on all 35 frames (recipe 0.315, two experts against two others 0.304) and left those on the 25 frames unchanged. A pair mean uses the readings present, and a frame enters a target's error only where both the reading and the pair mean exist. Because a pair mean falls back to one rater on the 18 frames R5 did not read, every score was recomputed with strict pairs that require both raters (Table~\ref{tab:strict}). On the 9 frames with no detected match, 4 were read by all seven raters and 5 by six. Strict pairs lowered the human references more than the readings: on the 9 frames the two-expert reference fell from 0.218 to 0.196 while the recipe moved from 0.213 to 0.209, so the recipe's point estimate went from 0.005 below the reference to 0.013 above it, inside an interval that still includes zero. On the 35 and the 25 frames the order of the readings did not change. In each resample the frames are drawn with replacement up to the size of the set, 5 of the 7 raters are drawn without replacement, and every score is recomputed over all pairs of the drawn raters. Resampling frames alone gives 53, 69 and 80\% of the interval width for the recipe minus the two-expert reference on the 9, the 35 and the 25 frames, and 76 to 93\% for the recipe minus DL\_Track\_US (\texttt{out/bench\_v4\_summary.csv}); the draw of raters gives the rest.

\begin{table}[H]
\centering
\begin{threeparttable}
\caption{Nearest training image of each benchmark frame. DL\_Track\_US set: four measures against the 3553 DL\_Track\_US training images; ORB counts the inlier keypoints of a RANSAC homography on the full frame and on its central half; screen: flagged by the thumbnail correlation (limit 0.97). Challenge set: central ORB inliers against the challenge training images (\texttt{umud audit challenge train}). Match: tissue match in either set; matched frames had 164 to 688 central inliers with a DL\_Track\_US image and 71 to 688 with a challenge image, the other nine at most 25 and 11.}
\label{tab:nearcopy}
\footnotesize
\setlength{\tabcolsep}{4.5pt}
\begin{tabular}{l r r r r r c r c}
\toprule
& \multicolumn{6}{c}{DL\_Track\_US training set} & Challenge set & \\
\cmidrule(lr){2-7}\cmidrule(lr){8-8}
Frame & Correlation & dHash bits & SSIM & ORB, full & ORB, centre & Screen & ORB, centre & Match \\
\midrule
im\_01 & 0.806 & 11 & 0.916 & 1291 & 322 & no & 166 & yes \\
im\_02 & 0.788 & 40 & 0.731 & 196 & 25 & no & 11 & no \\
im\_03 & 0.700 & 55 & 0.586 & 99 & 10 & no & 8 & no \\
im\_04 & 0.801 & 6 & 0.934 & 1356 & 308 & no & 308 & yes \\
im\_05 & 0.801 & 11 & 0.916 & 1312 & 334 & no & 235 & yes \\
im\_06 & 0.920 & 8 & 0.858 & 1271 & 525 & no & 525 & yes \\
im\_07 & 1.000 & 0 & 1.000 & 1433 & 209 & yes & 209 & yes \\
im\_08 & 0.900 & 9 & 0.866 & 1233 & 359 & no & 359 & yes \\
im\_09 & 0.986 & 9 & 0.926 & 1249 & 688 & yes & 688 & yes \\
im\_10 & 0.967 & 13 & 0.944 & 1218 & 460 & no & 460 & yes \\
im\_11 & 0.807 & 10 & 0.921 & 1288 & 286 & no & 166 & yes \\
im\_12 & 0.978 & 8 & 0.948 & 1235 & 429 & yes & 429 & yes \\
im\_13 & 0.980 & 9 & 0.948 & 1250 & 445 & yes & 445 & yes \\
im\_14 & 0.645 & 43 & 0.720 & 124 & 10 & no & 7 & no \\
im\_15 & 0.753 & 6 & 0.934 & 1315 & 330 & no & 168 & yes \\
im\_16 & 0.801 & 12 & 0.919 & 1269 & 291 & no & 158 & yes \\
im\_17 & 0.607 & 53 & 0.681 & 117 & 0 & no & 0 & no \\
im\_18 & 0.791 & 6 & 0.928 & 1297 & 277 & no & 277 & yes \\
im\_19 & 0.987 & 7 & 0.949 & 1258 & 219 & yes & 219 & yes \\
im\_20 & 0.821 & 17 & 0.892 & 872 & 177 & no & 71 & yes \\
im\_21 & 0.803 & 11 & 0.924 & 1311 & 384 & no & 230 & yes \\
im\_22 & 0.986 & 7 & 0.948 & 1286 & 347 & yes & 347 & yes \\
im\_23 & 0.336 & 89 & 0.261 & 8 & 5 & no & 5 & no \\
im\_24 & 0.328 & 91 & 0.259 & 9 & 6 & no & 7 & no \\
im\_25 & 0.978 & 8 & 0.939 & 1217 & 331 & yes & 331 & yes \\
im\_26 & 0.983 & 7 & 0.948 & 1222 & 387 & yes & 387 & yes \\
im\_27 & 0.776 & 12 & 0.921 & 1254 & 299 & no & 147 & yes \\
im\_28 & 0.985 & 8 & 0.949 & 1285 & 300 & yes & 300 & yes \\
im\_29 & 0.783 & 11 & 0.915 & 1324 & 422 & no & 221 & yes \\
im\_30 & 0.980 & 8 & 0.948 & 1230 & 531 & yes & 531 & yes \\
im\_31 & 0.340 & 89 & 0.257 & 10 & 6 & no & 7 & no \\
im\_32 & 0.808 & 6 & 0.929 & 1137 & 164 & no & 120 & yes \\
im\_33 & 0.761 & 34 & 0.716 & 127 & 9 & no & 6 & no \\
im\_34 & 0.723 & 52 & 0.609 & 134 & 9 & no & 6 & no \\
im\_35 & 0.829 & 12 & 0.920 & 1272 & 312 & no & 312 & yes \\
 
\bottomrule
\end{tabular}
\end{threeparttable}
\end{table}

\begin{table}[H]
\centering
\caption{Benchmark score against two-rater means as in Table~\ref{tab:bench}, but with strict pairs: a pair mean exists only where both raters read the frame. Intervals in brackets from 2000 resamples of frames and of 5 of the 7 raters (seed 2026).}
\label{tab:strict}
\footnotesize
\setlength{\tabcolsep}{4.5pt}
\begin{tabular}{l l l l}
\toprule
& All frames & Thumbnail-screen & No detected \\
Reading & ($n = 35$) & negative ($n = 25$) & tissue match ($n = 9$) \\
\midrule
Submission recipe & 0.276 [0.222, 0.333] & 0.257 [0.207, 0.320] & 0.209 [0.164, 0.262] \\
Leaderboard-free (post hoc) & 0.249 [0.204, 0.300] & 0.245 [0.198, 0.302] & 0.218 [0.165, 0.288] \\
DL\_Track\_US (sheet) & 0.336 [0.268, 0.412] & 0.339 [0.259, 0.447] & 0.259 [0.187, 0.333] \\
SMA (sheet) & 0.424 [0.283, 0.625] & 0.414 [0.260, 0.663] & 0.396 [0.289, 0.522] \\
One expert vs two others & 0.277 [0.232, 0.333] & 0.265 [0.215, 0.325] & 0.239 [0.189, 0.294] \\
Two experts vs two others & 0.227 [0.191, 0.273] & 0.217 [0.174, 0.266] & 0.196 [0.144, 0.241] \\
\midrule
Recipe minus DL\_Track\_US & $-0.060$ [$-$0.159, 0.027] & $-0.082$ [$-$0.187, 0.014] & $-0.050$ [$-$0.142, 0.038] \\
Recipe minus two experts & $+0.049$ [$-$0.003, 0.110] & $+0.040$ [$-$0.006, 0.095] & $+0.013$ [$-$0.048, 0.086] \\
Leaderboard-free minus two experts & $+0.022$ [$-$0.024, 0.076] & $+0.028$ [$-$0.019, 0.088] & $+0.021$ [$-$0.045, 0.101] \\
 
\bottomrule
\end{tabular}
\end{table}

\section{Blend sources, recordings and quality flags}
\label{app:blend}

\begin{figure}[H]
\centering
\includegraphics[width=\linewidth]{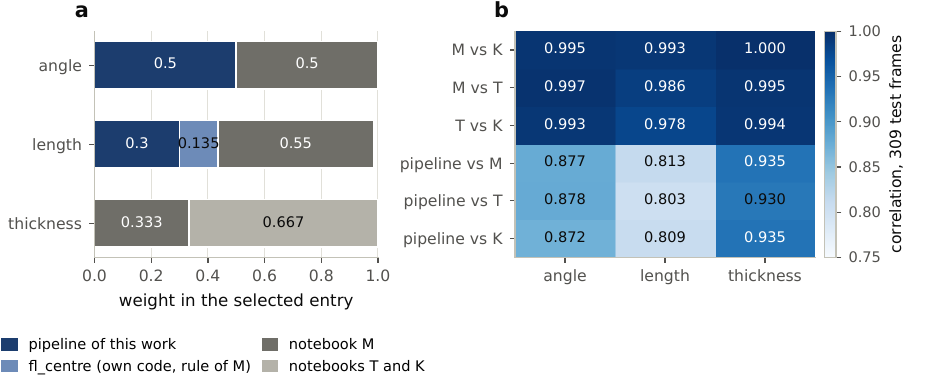}
\caption{Sources of the selected entry. (a)~Weight of each source in the final angle, length and thickness (offsets not shown). (b)~Correlation between the prediction files of the pipeline and the three public notebooks on the 309 test frames.}
\label{fig:sources}
\end{figure}

Within the 28 five-frame recordings the frames' own readings varied little before the median step: the median within-recording standard deviation was 0.12\textdegree, 0.84~mm in length and 0.008~mm in thickness, against between-recording standard deviations of 4.6\textdegree, 30.7~mm and 3.9~mm. The median moved a frame's angle by 0.15\textdegree{} and its length by 0.90~mm on average, with single outliers of 3.7\textdegree{} and 18.5~mm, and it differed from the recording mean by 0.09\textdegree{} and 0.46~mm on average. On the test set the flag rules would mark 4 frames as failed, 61 as low-confidence and 244 as valid. On the benchmark, a large error means any reading off the panel mean by more than half its tolerance; the flags gave 3 true and 5 false positives, 7 false and 20 true negatives, so they are auxiliary signs and not a basis for rejecting readings automatically.

\section{Replication details}
\label{app:replication}

Resampling the base entries within subsets gave wide intervals: sign agreement of 38 to 50\% for the corrections (4 bases), 61 to 93\% for the pipeline (7 bases) and 67 to 100\% for the blend (4 bases), and Pearson correlations of $-0.03$ to 0.45 for the corrections and $-0.48$ to 0.75 for angle changes (8 bases). Leaving out one base kept the overall correlation between 0.73 and 0.87. The 65 contrasts (Table~\ref{tab:effects}) come from 13 base entries: \entry{y\_full}, \entry{y\_b100}, \entry{y\_trust}, \entry{v2\_base\_filled}, \entry{b\_c88}, \entry{h\_v2\_mt}, notebook M, \entry{e\_pa50\_fl25}, \entry{f\_best}, \entry{h\_best}, \entry{k\_fl\_up3}, \entry{n\_comb} and \entry{p\_final}. Two were never submitted and were derived by additivity, by whole columns. \entry{f\_best} has angle 0.5 pipeline and 0.5 notebook M, length 0.35 pipeline and 0.65 notebook M, and thickness the mean of the three public files; \entry{e\_pa50\_fl35} differs from it only in thickness, which is notebook M's, and \entry{e\_pa50\_fl50\_mt3} only in length. During the competition, before \entry{h\_best} was submitted, the public score of \entry{f\_best} followed as \entry{e\_pa50\_fl35} $+$ \entry{e\_pa50\_fl50\_mt3} $-$ \entry{m\_mean\_pa} $-$ \entry{m\_mean\_fl} $+$ notebook M, 0.27009, and that of \entry{h\_best}, the blend with the angle and length changes of rows 41 and 45, as \entry{g\_pa\_up} $+$ \entry{g\_fl\_c} $-$ \entry{f\_best}, 0.24842. \entry{h\_best} served as the working base from 10 November on this derived score and was submitted late on 15 November to record its private score (0.24843 public). After that, the same identity read backwards, \entry{f\_best} $=$ \entry{g\_pa\_up} $+$ \entry{g\_fl\_c} $-$ \entry{h\_best}, gives 0.27008 public and 0.29373 private, and the private score of notebook M follows as \entry{m\_mean\_pa} $+$ \entry{m\_mean\_fl} $-$ \entry{e\_pa50\_fl50\_mt3} $+$ \entry{f\_best} $-$ \entry{e\_pa50\_fl35}, 0.34042; the same combination gives 0.30651 public against the published 0.30652, a difference of rounding. \entry{n\_comb}, four changes of \entry{h\_best} taken together, equals \entry{l\_fl\_ls3} $+$ \entry{m\_fl\_cg6} $-$ \entry{h\_best} $+$ \entry{l\_mt\_up1} $-$ \entry{k\_fl\_up3}: 0.24152 public and 0.27993 private with the recorded scores, the scores it received; the prediction made before \entry{h\_best} was submitted, with its derived public score, was 0.24153.

\begingroup
\footnotesize
\setlength{\tabcolsep}{3.5pt}
\begin{longtable}{>{\raggedright\arraybackslash}p{4.6cm} l l l r r c}
\caption{All 65 single-change contrasts: change, the submitted file, its base, the target, and the effect on the public and the private score (negative values are improvements). File names are given with underscores.}\label{tab:effects}\\
\toprule
Change & File & Base & Target & Public & Private & Same \\
\midrule
\endfirsthead
\toprule
Change & File & Base & Target & Public & Private & Same \\
\midrule
\endhead
\bottomrule
\endfoot
\multicolumn{7}{l}{\textit{Own pipeline}} \\
Telemed 800 frames read as square & \texttt{y\_telemed\_square} & \texttt{y\_full} & pixel shape & $+0.01277$ & $-0.00152$ & no \\
every frame read as square & \texttt{y\_scale\_only} & \texttt{y\_full} & pixel shape & $+0.02888$ & $+0.03688$ & yes \\
Lumify angle read as square & \texttt{i\_lum\_pa\_sq} & \texttt{h\_best} & pixel shape & $+0.02054$ & $+0.01693$ & yes \\
Lumify length read as square & \texttt{i\_lum\_fl\_sq} & \texttt{h\_best} & pixel shape & $+0.02637$ & $+0.01778$ & yes \\
length marched along the fascicle & \texttt{y\_full\_marched} & \texttt{y\_full} & length & $+0.06592$ & $+0.11726$ & yes \\
closed form at fixed weight 0.65 & \texttt{y\_full\_c65} & \texttt{y\_full} & length & $-0.03083$ & $-0.01821$ & yes \\
closed form at fixed weight 0.8 & \texttt{y\_b80} & \texttt{y\_b100} & length & $-0.05131$ & $-0.04823$ & yes \\
trust weight, k = 0.176 & \texttt{y\_trust} & \texttt{y\_b100} & length & $-0.05749$ & $-0.06737$ & yes \\
k = 0.23 instead of 0.176 & \texttt{y\_trust\_k23} & \texttt{y\_trust} & length & $+0.01085$ & $+0.00054$ & yes \\
k = 0.13 instead of 0.176 & \texttt{y\_trust\_k13} & \texttt{y\_trust} & length & $+0.01108$ & $+0.01373$ & yes \\
centre 88 mm instead of 83 & \texttt{b\_c88} & \texttt{h\_v2\_mt} & length & $-0.00842$ & $-0.00099$ & yes \\
centre 78 mm instead of 83 & \texttt{b\_c78} & \texttt{h\_v2\_mt} & length & $+0.01167$ & $+0.00671$ & yes \\
centre 98 mm instead of 83 & \texttt{b\_c98} & \texttt{h\_v2\_mt} & length & $-0.00775$ & $+0.00960$ & no \\
centre 93 mm instead of 83 & \texttt{b\_c93} & \texttt{h\_v2\_mt} & length & $-0.01065$ & $+0.00216$ & no \\
centre at each layout median & \texttt{b\_fam} & \texttt{h\_v2\_mt} & length & $+0.02566$ & $+0.02289$ & yes \\
centre halfway to layout median & \texttt{b\_fam\_half} & \texttt{h\_v2\_mt} & length & $+0.00464$ & $+0.00633$ & yes \\
k = 0.15 at centre 88 mm & \texttt{b\_c88\_k15} & \texttt{b\_c88} & length & $+0.00629$ & $+0.00695$ & yes \\
angle from five-fold fascicle models & \texttt{h\_v2\_pa} & \texttt{y\_trust} & angle & $+0.01364$ & $-0.01205$ & no \\
thickness from five-fold aponeurosis models & \texttt{h\_v2\_mt} & \texttt{y\_trust} & thickness & $-0.00777$ & $-0.00548$ & yes \\
mean of old and five-fold angles & \texttt{h\_mean\_pa} & \texttt{y\_trust} & angle & $+0.01007$ & $-0.00925$ & no \\
angle from v3 fascicle models & \texttt{h\_v3\_pa} & \texttt{h\_v2\_mt} & angle & $+0.00935$ & $-0.00508$ & no \\
angle at the deep insertion & \texttt{v2\_deep\_filled} & \texttt{v2\_base\_filled} & angle & $+0.00203$ & $+0.03363$ & yes \\
angle +0.6\textdegree{} & \texttt{b\_pa\_up} & \texttt{h\_v2\_mt} & angle & $+0.01593$ & $+0.00565$ & yes \\
angle $-$0.6\textdegree{} & \texttt{b\_pa\_down} & \texttt{h\_v2\_mt} & angle & $-0.00063$ & $+0.01143$ & no \\
\multicolumn{7}{l}{\textit{Blend}} \\
M with our angle & \texttt{m\_our\_pa} & \texttt{notebook M} & angle & $-0.00243$ & $-0.00281$ & yes \\
M with our thickness & \texttt{m\_our\_mt} & \texttt{notebook M} & thickness & $+0.03845$ & $+0.02434$ & yes \\
angle: mean of M and ours & \texttt{m\_mean\_pa} & \texttt{notebook M} & angle & $-0.01298$ & $-0.01354$ & yes \\
thickness: mean of M and ours & \texttt{m\_mean\_mt} & \texttt{notebook M} & thickness & $+0.01601$ & $+0.00944$ & yes \\
length: mean of M and ours & \texttt{m\_mean\_fl} & \texttt{notebook M} & length & $-0.01654$ & $-0.02625$ & yes \\
length 0.35 ours instead of 0.25 & \texttt{e\_pa50\_fl35} & \texttt{e\_pa50\_fl25} & length & $-0.00179$ & $-0.00294$ & yes \\
angle half ours, half mean of public files & \texttt{f\_pa\_pub3} & \texttt{f\_best} & angle & $+0.00148$ & $+0.00030$ & yes \\
thickness: mean of public and our centre-line & \texttt{f\_mt\_mean\_ch} & \texttt{f\_best} & thickness & $+0.00764$ & $+0.00550$ & yes \\
thickness: our centre-line reading & \texttt{f\_mt\_ch} & \texttt{f\_best} & thickness & $+0.02053$ & $+0.01647$ & yes \\
closed form from blend at 0.5 & \texttt{f\_fl\_ens50} & \texttt{f\_best} & length & $+0.01419$ & $+0.01411$ & yes \\
closed form from blend at 0.35 & \texttt{f\_fl\_ens35} & \texttt{f\_best} & length & $+0.00882$ & $+0.01242$ & yes \\
blend angle +0.4\textdegree{} & \texttt{g\_pa\_up} & \texttt{f\_best} & angle & $-0.00341$ & $-0.00850$ & yes \\
blend angle $-$0.4\textdegree{} & \texttt{g\_pa\_down} & \texttt{f\_best} & angle & $+0.00824$ & $+0.01175$ & yes \\
thickness +0.25 mm & \texttt{g\_mt\_up} & \texttt{f\_best} & thickness & $+0.00089$ & $+0.00333$ & yes \\
thickness $-$0.25 mm & \texttt{g\_mt\_down} & \texttt{f\_best} & thickness & $+0.01658$ & $+0.01643$ & yes \\
fl\_centre at 0.15 & \texttt{g\_fl\_c} & \texttt{f\_best} & length & $-0.01824$ & $-0.00940$ & yes \\
fl\_lin3 at 0.15 instead of fl\_centre & \texttt{j\_fl\_lin15} & \texttt{h\_best} & length & $+0.01276$ & $-0.00146$ & no \\
fl\_lin3 at 0.30 instead of fl\_centre & \texttt{j\_fl\_lin30} & \texttt{h\_best} & length & $+0.01321$ & $-0.00331$ & no \\
fl\_lin3 at 0.45 instead of fl\_centre & \texttt{j\_fl\_lin45} & \texttt{h\_best} & length & $+0.02005$ & $+0.00155$ & yes \\
fl\_centre at 0.30 & \texttt{h\_fl\_c30} & \texttt{h\_best} & length & $+0.00446$ & $+0.01123$ & yes \\
fl\_centre at 0.45 & \texttt{h\_fl\_c45} & \texttt{h\_best} & length & $+0.01746$ & $+0.03351$ & yes \\
\multicolumn{7}{l}{\textit{Late-stage corrections}} \\
all lengths $\times$1.03 & \texttt{k\_fl\_up3} & \texttt{h\_best} & length & $-0.00248$ & $+0.01711$ & no \\
all lengths $\times$0.97 & \texttt{k\_fl\_dn3} & \texttt{h\_best} & length & $+0.02184$ & $+0.00605$ & yes \\
all lengths $\times$1.015 & \texttt{l\_fl\_up15} & \texttt{h\_best} & length & $-0.00406$ & $+0.00613$ & no \\
all lengths $\times$1.045 & \texttt{l\_fl\_up45} & \texttt{h\_best} & length & $+0.00367$ & $+0.03308$ & yes \\
Telemed lengths $\times$1.03 & \texttt{l\_fl\_tm3} & \texttt{h\_best} & length & $+0.00208$ & $+0.01223$ & yes \\
Telemed lengths $\times$0.98 & \texttt{m\_fl\_tm\_dn2} & \texttt{h\_best} & length & $+0.00496$ & $-0.00066$ & no \\
Siemens and Lumify lengths $\times$1.03 & \texttt{l\_fl\_ls3} & \texttt{h\_best} & length & $-0.00204$ & $+0.00678$ & no \\
Siemens and Lumify lengths $\times$1.015 & \texttt{m\_fl\_ls15} & \texttt{h\_best} & length & $-0.00137$ & $+0.00263$ & no \\
Siemens and Lumify lengths $\times$1.05 & \texttt{m\_fl\_ls5} & \texttt{h\_best} & length & $+0.00033$ & $+0.01411$ & yes \\
crop and grey lengths $\times$1.06 & \texttt{m\_fl\_cg6} & \texttt{h\_best} & length & $-0.00264$ & $-0.00158$ & yes \\
pa\_ins3 at 0.15 in the angle & \texttt{k\_pa\_ins} & \texttt{h\_best} & angle & $+0.00044$ & $-0.00181$ & no \\
thickness +0.1 mm & \texttt{l\_mt\_up1} & \texttt{k\_fl\_up3} & thickness & $-0.00223$ & $-0.00110$ & yes \\
thickness +0.2 mm & \texttt{m\_mt\_up2} & \texttt{k\_fl\_up3} & thickness & $-0.00086$ & $+0.00098$ & no \\
Telemed 644 lengths $\times$1.03 & \texttt{n\_t644\_up3} & \texttt{n\_comb} & length & $+0.00079$ & $+0.00183$ & yes \\
Siemens lengths back to $\times$1 & \texttt{n\_sie\_back} & \texttt{n\_comb} & length & $+0.00423$ & $-0.00276$ & no \\
Telemed angles +0.5\textdegree{} & \texttt{n\_pa\_tm\_up} & \texttt{n\_comb} & angle & $-0.00230$ & $-0.00385$ & yes \\
Siemens and Lumify angles +0.5\textdegree{} & \texttt{n\_pa\_ls\_up} & \texttt{n\_comb} & angle & $+0.00148$ & $-0.00126$ & no \\
Lumify lengths $\times$0.97 & \texttt{p\_final\_lum\_dn3} & \texttt{p\_final} & length & $-0.00020$ & $-0.00220$ & yes \\
Telemed angles +1.0\textdegree{} instead of +0.5\textdegree{} & \texttt{p\_final\_pa\_tm10} & \texttt{p\_final} & angle & $+0.00008$ & $-0.00019$ & no \\
Siemens lengths $\times$1.05 instead of $\times$1.03 & \texttt{p\_final\_sie5} & \texttt{p\_final} & length & $-0.00003$ & $+0.00350$ & no \\
 
\end{longtable}
\endgroup

\begin{table}[H]
\centering
\begin{threeparttable}
\caption{Simulated effect of recordings on the noise estimate. Factor: spread of a public effect relative to the value of Eq.~\eqref{eq:se} for independent frames; noise correlation: correlation between public and private effects when the true effect is zero. $\rho$: share of the variance of a frame's change that is common to its recording.}
\label{tab:splitsim}
\small
\begin{tabular}{c c c c c}
\toprule
& \multicolumn{2}{c}{Random frame split} & \multicolumn{2}{c}{Split by recording} \\
$\rho$ & Factor & Noise correlation & Factor & Noise correlation \\
\midrule
0.00 & 1.01 & 0.00 & 0.98 & $-0.02$ \\
0.25 & 1.09 & $+0.20$ & 1.00 & 0.00 \\
0.50 & 1.14 & $+0.31$ & 0.99 & 0.00 \\
0.75 & 1.22 & $+0.40$ & 1.01 & $+0.02$ \\
0.90 & 1.26 & $+0.44$ & 1.01 & $+0.02$ \\
 
\bottomrule
\end{tabular}
\end{threeparttable}
\end{table}

\begin{figure}[H]
\centering
\includegraphics[width=\linewidth]{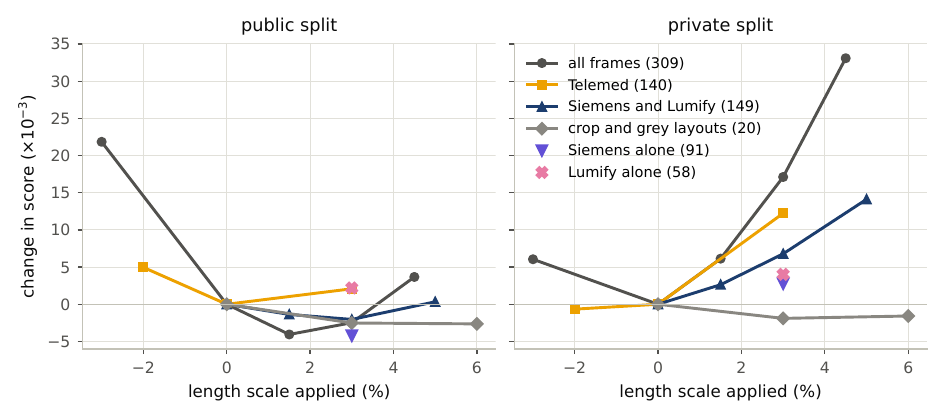}
\caption{Change in score against the length scale applied to one group of layouts, everything else as in \entry{h\_best}, on the public split (left) and the private split (right). The crop and grey value at 3\% is derived from the other groups. Siemens and Lumify, measured together, pull in opposite directions on the public split.}
\label{fig:scale}
\end{figure}

\section{Recommendations}
\label{app:recs}

\begin{table}[H]
\centering
\caption{Recommendations that follow from the evidence of this study.}
\label{tab:recs}
\small
\begin{tabular}{>{\raggedright\arraybackslash}p{2.1cm} >{\raggedright\arraybackslash}p{5.9cm} >{\raggedright\arraybackslash}p{6.5cm}}
\toprule
For & Evidence in this study & Recommendation \\
\midrule
Datasets & square-pixel reading cost 0.037 private; a 2\% scale error exceeds rater spread in thickness & store the pixel spacing along both axes with every frame; state how frames were resized for release \\
Datasets & extension conventions differ by a median of 21~mm below 12\textdegree & publish the extension rule and the thickness direction with the labels \\
Benchmarks & 26 of 35 frames share tissue with two published training sets & check tissue-level overlap with public training sets, with matching restricted to the image region and in both left-right orientations \\
Challenges & RMS-based dispersion scale of a public effect 0.005 to 0.008 on 103 frames (this entry) & report a noise estimate with the leaderboard, describe how frames were split, and split by recording \\
Challenges & 9 of 20 small corrections kept their sign & limit or delay leaderboard feedback for fine tuning \citep{blum2015}; report private effects of decisions \\
Tools & flags caught 3 of 10 large errors; four frames received a silent fallback & write a status, a fallback type and an uncertainty validated on independent data with every reading, and allow a tool to withhold a length \\
\bottomrule
\end{tabular}
\end{table}

\section{Scored submissions}
\label{app:subs}

\begin{figure}[H]
\centering
\includegraphics[width=\linewidth]{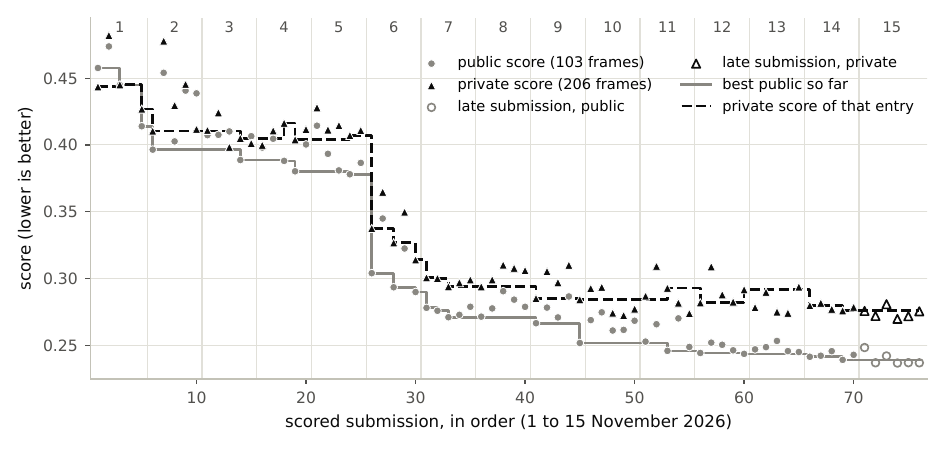}
\caption{Public (circles) and private (triangles) score of every scored submission from 1 to 15 November 2026, in order of scoring; numbers along the top are days of November, and open symbols are late submissions. The solid step follows the best public score so far and the dashed step the private score of that same entry. Row 4 (0.51076 public, 0.56239 private) and the entries before 1 November lie above the scale.}
\label{fig:lb}
\end{figure}

Table~\ref{tab:subs} lists the 76 scored submissions from 1 to 15 November 2026 in the order in which scores were received: rows 1 to 70 were scored before the deadline and rows 71 to 76, marked with asterisks, after it. Public scores are on 103 frames and private scores on 206, read from the Kaggle submissions page after the private leaderboard was released. The selected entries are rows 60 and 69. Rows 26 to 30 change one column of notebook M. Rows 31 to 35 blend notebook M with the pipeline at other weights and keep notebook M's thickness, except row 32, which takes the mean thickness of the three public files. Rows 36 to 45 change one column of \entry{f\_best} (rows 41 to 45 carry the prefix g), and rows 46 to 64, except row 60, one column of \entry{h\_best}; rows 60 and 65 add a thickness offset to \entry{k\_fl\_up3} (row 53). Row 66, \entry{n\_comb}, combines four changes of \entry{h\_best}, and rows 67 to 70 change one column of it. The derived bases are given in Section~\ref{app:replication}. Table~\ref{tab:earlier} lists the 48 earlier entries, made before 1 November with a pipeline that read a ruler on every frame; with rows 1 to 70 they are the 118 entries shown for the team. Figure~\ref{fig:lb} shows the sequence of scores.

\begingroup
\footnotesize
\begin{longtable}{r l l r r >{\raggedright\arraybackslash}p{6.3cm}}
\caption{All scored submissions from 1 to 15 November 2026. Asterisks mark late submissions.}\label{tab:subs}\\
\toprule
\# & Date & File & Public & Private & What changed \\
\midrule
\endfirsthead
\toprule
\# & Date & File & Public & Private & What changed \\
\midrule
\endhead
\bottomrule
\endfoot
1 & 01 Nov & \texttt{y\_telemed\_square} & 0.45761 & 0.44361 & layout scales and pixel aspect, Telemed 800 $\times$ 1200 frames read as square \\
2 & 01 Nov & \texttt{y\_scale\_only} & 0.47372 & 0.48201 & layout scales, every frame read as square \\
3 & 01 Nov & \texttt{y\_full} & 0.44484 & 0.44513 & layout scales and pixel aspect for every layout \\
4 & 01 Nov & \texttt{y\_full\_marched} & 0.51076 & 0.56239 & as y\_full, length marched along the fascicle \\
5 & 01 Nov & \texttt{y\_full\_c65} & 0.41401 & 0.42692 & as y\_full, closed-form length at weight 0.65 towards 83 mm \\
6 & 02 Nov & \texttt{y\_trust} & 0.39650 & 0.41048 & per-frame weight $1/(1+(0.176\cot\mathrm{PA})^2)$ towards 83 mm \\
7 & 02 Nov & \texttt{y\_b100} & 0.45399 & 0.47785 & closed-form length alone \\
8 & 02 Nov & \texttt{y\_b80} & 0.40268 & 0.42962 & closed-form length at weight 0.8 \\
9 & 02 Nov & \texttt{v2\_deep\_filled} & 0.44066 & 0.44544 & v2 models, angle at the deep insertion \\
10 & 02 Nov & \texttt{v2\_base\_filled} & 0.43863 & 0.41181 & v2 models, fragment angle \\
11 & 03 Nov & \texttt{y\_trust\_k23} & 0.40735 & 0.41102 & y\_trust with k = 0.23 \\
12 & 03 Nov & \texttt{y\_trust\_k13} & 0.40758 & 0.42421 & y\_trust with k = 0.13 \\
13 & 03 Nov & \texttt{h\_v2\_pa} & 0.41014 & 0.39843 & y\_trust with the v2 angle \\
14 & 03 Nov & \texttt{h\_v2\_mt} & 0.38873 & 0.40500 & y\_trust with the v2 thickness \\
15 & 03 Nov & \texttt{h\_mean\_pa} & 0.40657 & 0.40123 & y\_trust with the mean of both angles \\
16 & 04 Nov & \texttt{h\_v3\_pa} & 0.39808 & 0.39992 & h\_v2\_mt with the v3 angle \\
17 & 04 Nov & \texttt{b\_pa\_up} & 0.40466 & 0.41065 & h\_v2\_mt, angle $+$0.6\textdegree{} \\
18 & 04 Nov & \texttt{b\_pa\_down} & 0.38810 & 0.41643 & h\_v2\_mt, angle $-$0.6\textdegree{} \\
19 & 04 Nov & \texttt{b\_c88} & 0.38031 & 0.40401 & h\_v2\_mt, length centre 88 mm \\
20 & 04 Nov & \texttt{b\_c78} & 0.40040 & 0.41171 & h\_v2\_mt, length centre 78 mm \\
21 & 05 Nov & \texttt{b\_fam} & 0.41439 & 0.42789 & length centre at each layout's median \\
22 & 05 Nov & \texttt{b\_fam\_half} & 0.39337 & 0.41133 & centre halfway between 83 mm and the layout median \\
23 & 05 Nov & \texttt{b\_c98} & 0.38098 & 0.41460 & length centre 98 mm \\
24 & 05 Nov & \texttt{b\_c93} & 0.37808 & 0.40716 & length centre 93 mm; best entry of the pipeline alone \\
25 & 05 Nov & \texttt{b\_c88\_k15} & 0.38660 & 0.41096 & centre 88 mm, k = 0.15 \\
26 & 06 Nov & \texttt{m\_our\_pa} & 0.30409 & 0.33761 & public two U-Net file with the pipeline's angle \\
27 & 06 Nov & \texttt{m\_our\_mt} & 0.34497 & 0.36476 & public two U-Net file with the pipeline's thickness \\
28 & 06 Nov & \texttt{m\_mean\_pa} & 0.29354 & 0.32688 & angle: mean of the two \\
29 & 06 Nov & \texttt{m\_mean\_mt} & 0.32253 & 0.34986 & thickness: mean of the two \\
30 & 06 Nov & \texttt{m\_mean\_fl} & 0.28998 & 0.31417 & length: mean of the two \\
31 & 07 Nov & \texttt{e\_pa65\_fl50} & 0.27814 & 0.30084 & angle 0.65 own, length 0.5 own, thickness of notebook M \\
32 & 07 Nov & \texttt{e\_pa50\_fl50\_mt3} & 0.27596 & 0.30025 & angle and length 0.5 own, thickness mean of three public files \\
33 & 07 Nov & \texttt{e\_pa50\_fl35} & 0.27113 & 0.29411 & angle 0.5 own, length 0.35 own, thickness of notebook M \\
34 & 07 Nov & \texttt{e\_pa50\_fl25} & 0.27292 & 0.29705 & angle 0.5 own, length 0.25 own, thickness of notebook M \\
35 & 07 Nov & \texttt{e\_pa3\_fl50} & 0.27895 & 0.29921 & angle: mean of public, own and the pipeline's texture angle, thickness of notebook M \\
36 & 08 Nov & \texttt{f\_pa\_pub3} & 0.27156 & 0.29403 & angle half own, half the mean of three public files \\
37 & 08 Nov & \texttt{f\_mt\_mean\_ch} & 0.27772 & 0.29923 & thickness: mean of public and the pipeline's band-centre reading \\
38 & 08 Nov & \texttt{f\_mt\_ch} & 0.29061 & 0.31020 & thickness: the pipeline's band-centre reading \\
39 & 08 Nov & \texttt{f\_fl\_ens50} & 0.28427 & 0.30784 & length 0.5 public, 0.5 closed form from the blended angle and thickness \\
40 & 08 Nov & \texttt{f\_fl\_ens35} & 0.27890 & 0.30615 & the same at 0.35 \\
41 & 09 Nov & \texttt{g\_pa\_up} & 0.26667 & 0.28523 & angle $+$0.4\textdegree{} \\
42 & 09 Nov & \texttt{g\_pa\_down} & 0.27832 & 0.30548 & angle $-$0.4\textdegree{} \\
43 & 09 Nov & \texttt{g\_mt\_up} & 0.27097 & 0.29706 & thickness $+$0.25 mm \\
44 & 09 Nov & \texttt{g\_mt\_down} & 0.28666 & 0.31016 & thickness $-$0.25 mm \\
45 & 09 Nov & \texttt{g\_fl\_c} & 0.25184 & 0.28433 & length 0.55 public, 0.30 own, 0.15 $\times$ 0.9 $\times$ fl\_centre \\
46 & 10 Nov & \texttt{i\_lum\_pa\_sq} & 0.26897 & 0.29276 & Lumify angle as read with square pixels \\
47 & 10 Nov & \texttt{i\_lum\_fl\_sq} & 0.27480 & 0.29361 & Lumify length as read with square pixels \\
48 & 10 Nov & \texttt{j\_fl\_lin15} & 0.26119 & 0.27437 & fl\_lin3 at weight 0.15 in place of fl\_centre \\
49 & 10 Nov & \texttt{j\_fl\_lin30} & 0.26164 & 0.27252 & fl\_lin3 at weight 0.30 \\
50 & 10 Nov & \texttt{j\_fl\_lin45} & 0.26848 & 0.27738 & fl\_lin3 at weight 0.45 \\
51 & 11 Nov & \texttt{h\_fl\_c30} & 0.25289 & 0.28706 & fl\_centre at weight 0.30 \\
52 & 11 Nov & \texttt{h\_fl\_c45} & 0.26589 & 0.30934 & fl\_centre at weight 0.45 \\
53 & 11 Nov & \texttt{k\_fl\_up3} & 0.24595 & 0.29294 & all lengths $\times$ 1.03 \\
54 & 11 Nov & \texttt{k\_fl\_dn3} & 0.27027 & 0.28188 & all lengths $\times$ 0.97 \\
55 & 11 Nov & \texttt{k\_pa\_ins} & 0.24887 & 0.27402 & pa\_ins3 at weight 0.15 in the angle \\
56 & 12 Nov & \texttt{l\_fl\_up15} & 0.24437 & 0.28196 & all lengths $\times$ 1.015 \\
57 & 12 Nov & \texttt{l\_fl\_up45} & 0.25210 & 0.30891 & all lengths $\times$ 1.045 \\
58 & 12 Nov & \texttt{l\_fl\_tm3} & 0.25051 & 0.28806 & Telemed lengths $\times$ 1.03 \\
59 & 12 Nov & \texttt{l\_fl\_ls3} & 0.24639 & 0.28261 & Siemens and Lumify lengths $\times$ 1.03 \\
60 & 12 Nov & \texttt{l\_mt\_up1} & 0.24372 & 0.29184 & k\_fl\_up3 with thickness $+$0.1 mm \\
61 & 13 Nov & \texttt{m\_fl\_ls15} & 0.24706 & 0.27846 & Siemens and Lumify lengths $\times$ 1.015 \\
62 & 13 Nov & \texttt{m\_fl\_ls5} & 0.24876 & 0.28994 & Siemens and Lumify lengths $\times$ 1.05 \\
63 & 13 Nov & \texttt{m\_fl\_tm\_dn2} & 0.25339 & 0.27517 & Telemed lengths $\times$ 0.98 \\
64 & 13 Nov & \texttt{m\_fl\_cg6} & 0.24579 & 0.27425 & crop and grey layouts lengths $\times$ 1.06 \\
65 & 13 Nov & \texttt{m\_mt\_up2} & 0.24509 & 0.29392 & k\_fl\_up3 with thickness $+$0.2 mm \\
66 & 14 Nov & \texttt{n\_comb} & 0.24152 & 0.27993 & Siemens and Lumify $\times$ 1.03, crop and grey $\times$ 1.06, thickness $+$0.1 mm \\
67 & 14 Nov & \texttt{n\_t644\_up3} & 0.24231 & 0.28176 & n\_comb with Telemed 644 lengths $\times$ 1.03 \\
68 & 14 Nov & \texttt{n\_sie\_back} & 0.24575 & 0.27717 & n\_comb with Siemens lengths back to $\times$ 1 \\
69 & 14 Nov & \texttt{n\_pa\_tm\_up} & 0.23922 & 0.27608 & n\_comb with Telemed angle $+$0.5\textdegree{} \\
70 & 14 Nov & \texttt{n\_pa\_ls\_up} & 0.24300 & 0.27867 & n\_comb with Siemens and Lumify angle $+$0.5\textdegree{} \\
71 & 15 Nov & \texttt{h\_best} & 0.24843* & 0.27583 & g\_fl\_c with angle $+$0.4\textdegree{} \\
72 & 15 Nov & \texttt{p\_final} & 0.23704* & 0.27206 & n\_pa\_tm\_up with Lumify lengths back to $\times$ 1 \\
73 & 15 Nov & \texttt{p\_safe} & 0.24214* & 0.28086 & all lengths $\times$ 1.015, thickness $+$0.1 mm, no per-layout terms \\
74 & 15 Nov & \texttt{p\_final\_lum\_dn3} & 0.23684* & 0.26986 & p\_final with Lumify lengths $\times$ 0.97 \\
75 & 15 Nov & \texttt{p\_final\_pa\_tm10} & 0.23712* & 0.27187 & p\_final with Telemed angle $+$1.0\textdegree{} \\
76 & 15 Nov & \texttt{p\_final\_sie5} & 0.23701* & 0.27556 & p\_final with Siemens lengths $\times$ 1.05 \\
 
\end{longtable}
\endgroup

\begingroup
\footnotesize
\begin{longtable}{l l r r l}
\caption{Scored entries made before 1 November 2026 with the earlier pipeline, in order of scoring. Probe entries gave every frame one value or shifted one reading.}\label{tab:earlier}\\
\toprule
& File & Public & Private & Note \\
\midrule
\endfirsthead
\toprule
& File & Public & Private & Note \\
\midrule
\endhead
\bottomrule
\endfoot
E1 & \texttt{v\_new\_100} & 0.52995 & 0.57348 &  \\
E2 & \texttt{v\_hyb\_108} & 0.70341 & 0.75912 &  \\
E3 & \texttt{v\_hyb\_104} & 0.61653 & 0.66315 &  \\
E4 & \texttt{v\_hyb\_100} & 0.54142 & 0.58374 &  \\
E5 & \texttt{v\_hyb\_096} & 0.54426 & 0.58688 &  \\
E6 & \texttt{marched} & 0.65455 & 0.72843 &  \\
E7 & \texttt{thicker} & 0.54481 & 0.57999 &  \\
E8 & \texttt{best} & 0.59833 & 0.62816 &  \\
E9 & \texttt{u\_frag\_smoothed} & 0.83783 & 0.85238 &  \\
E10 & \texttt{u\_frag\_blend50} & 0.65483 & 0.68317 &  \\
E11 & \texttt{r\_mt088} & 0.87213 & 0.89890 &  \\
E12 & \texttt{r\_mt084} & 0.93562 & 0.97572 &  \\
E13 & \texttt{r\_newfrag} & 0.82966 & 0.82285 &  \\
E14 & \texttt{s\_fl\_closed} & 0.74421 & 0.84058 &  \\
E15 & \texttt{s\_fl\_blend} & 0.70177 & 0.76225 &  \\
E16 & \texttt{r\_fl90} & 0.86185 & 0.87015 &  \\
E17 & \texttt{r\_fl70} & 0.91469 & 0.96431 &  \\
E18 & \texttt{p\_old\_angle\_mt092} & 0.84457 & 0.85723 &  \\
E19 & \texttt{p\_old\_angle} & 0.88187 & 0.85239 &  \\
E20 & \texttt{p\_level} & 0.99120 & 0.93968 &  \\
E21 & \texttt{p\_level\_mt092} & 0.95391 & 0.94452 &  \\
E22 & \texttt{thin\_5} & 1.30191 & 1.29590 &  \\
E23 & \texttt{flat\_length} & 1.04809 & 0.98583 &  \\
E24 & \texttt{blend\_half} & 1.15265 & 1.11134 &  \\
E25 & \texttt{measured} & 1.33458 & 1.30358 &  \\
E26 & \texttt{x\_repaired\_fl\_measured} & 0.99535 & 1.07177 &  \\
E27 & \texttt{w\_repaired\_pa12\_mt092} & 0.92233 & 0.89891 &  \\
E28 & \texttt{v\_repaired\_mt092} & 0.93738 & 0.91992 &  \\
E29 & \texttt{u\_repaired\_pa12} & 1.00193 & 0.96548 &  \\
E30 & \texttt{t\_repaired} & 1.01698 & 0.98649 &  \\
E31 & \texttt{d\_mt\_070\_pa\_plus12} & 1.18591 & 1.22934 &  \\
E32 & \texttt{c\_mt\_090} & 1.02375 & 0.97447 &  \\
E33 & \texttt{a\_mt\_070} & 1.20096 & 1.25035 &  \\
E34 & \texttt{Kaggle Pipeline - Version 1} & 1.15311 & 1.07309 &  \\
E35 & \texttt{Kaggle Pipeline - Version 1} & 1.19650 & 1.12190 &  \\
E36 & \texttt{v\_pa\_plus3} & 0.96865 & 0.92949 &  \\
E37 & \texttt{v\_pa\_plus1\_2} & 0.93983 & 0.91251 &  \\
E38 & \texttt{v\_fl\_98} & 1.05789 & 1.02053 &  \\
E39 & \texttt{v\_fl\_62} & 1.11363 & 1.17260 &  \\
E40 & \texttt{sub\_models\_derived\_fl} & 0.98969 & 0.89383 &  \\
E41 & \texttt{sub\_models\_flat\_fl} & 0.95089 & 0.89793 &  \\
E42 & \texttt{probe\_mt\_const22} & 1.21315 & 1.20613 & constant or shifted probe \\
E43 & \texttt{probe\_fl\_plus20} & 1.25661 & 1.18240 & constant or shifted probe \\
E44 & \texttt{probe\_fl\_const80} & 1.04988 & 0.96533 & constant or shifted probe \\
E45 & \texttt{probe\_mt\_plus3} & 1.43021 & 1.40277 & constant or shifted probe \\
E46 & \texttt{probe\_pa\_23} & 1.39461 & 1.37244 & constant or shifted probe \\
E47 & \texttt{probe\_pa\_11} & 1.37388 & 1.35914 & constant or shifted probe \\
E48 & \texttt{sub\_a\_comma} & 1.19737 & 1.18037 &  \\
 
\end{longtable}
\endgroup

\section{Reproducibility}
\label{app:repro}

\paragraph{Levels of reproduction} Statistical reproduction rebuilds every number, table and figure of the paper from the derived files with the scripts in \texttt{analysis/}; it was carried out for this version. Submission reproduction rebuilds the blended files from \texttt{b c93.csv} and the three notebook outputs with \texttt{scripts/blend.py}; \entry{h\_best}, \entry{n\_pa\_tm\_up} and \entry{p\_final} come out byte-identical to the submitted files, and the second selected entry, \entry{l\_mt\_up1}, comes out byte-identical from \texttt{h best.csv} with \texttt{analysis/rebuild\_l\_mt\_up1.py}. Prediction reproduction was carried out on Kaggle on 24 November 2026: the notebook \texttt{umud reproduce predictions} ran \texttt{scripts/predict.py} of version 0.18.1 with the published weights on the 309 test frames, and both \texttt{raw\_predictions.csv} and \texttt{prediction\_report.csv} came out with the same SHA-256 as the files of the original run of 9 November (version 3 of \texttt{umud v6 predict}, code 0.18.0); all 20 reading columns agree exactly. Versions 0.18.0 and 0.18.1 differ only in files that take no part in prediction (the version string, \texttt{blend.py}, \texttt{score\_benchmark.py}, the added \texttt{check\_overlap.py}, the README and the package metadata). The pipeline file \entry{b\_c93} itself is a historical file, assembled from two runs, and is not the output of one execution of one configuration: its angle comes from an earlier run with the single-fold fascicle model under an earlier code version, rebuilt offline with the layout scales, and differs from the angle of the test-set run; the newer fascicle models had lost on the public split (Section~\ref{sec:res-replication}). Its thickness comes from the run with the five-fold aponeurosis models, which gained on the public split, and its length agrees within 0.8~mm with the recipe of Eq.~\eqref{eq:trust} computed from these two columns; the columns were joined by image identifier. No script that rebuilds it in one step was kept, so the file is published as it was submitted and identified by its hash.

\paragraph{Kaggle resources} Table~\ref{tab:kaggle} lists the author's notebooks and datasets. All are public on Kaggle, at the addresses given in the table. The weights were gathered from the outputs of the two training notebooks by \texttt{umud collect weights}, which records their hashes, and published unchanged.

\begin{table}[H]
\centering
\caption{The author's Kaggle notebooks and datasets; each name links to its page. Version: the Kaggle version used in this paper. Dates are run dates in 2026.}
\label{tab:kaggle}
\footnotesize
\begin{tabular}{l c >{\raggedright\arraybackslash}p{8.2cm}}
\toprule
Name & Version & Role \\
\midrule
\multicolumn{3}{l}{Notebooks, at \texttt{https://www.kaggle.com/code/soheilkooklan/} followed by the name} \\
\href{https://www.kaggle.com/code/soheilkooklan/train-fasc}{\texttt{train-fasc}} & 1 & fascicle model, single fold, code 0.12.0 \\
\href{https://www.kaggle.com/code/soheilkooklan/umud-v2-train-apo}{\texttt{umud-v2-train-apo}} & 1 & aponeurosis models, five folds, code 0.14.0, 1 November \\
\href{https://www.kaggle.com/code/soheilkooklan/umud-v3-predict}{\texttt{umud-v3-predict}} & 3 & earlier test run \\
\href{https://www.kaggle.com/code/soheilkooklan/umud-v4-predict}{\texttt{umud-v4-predict}} & 1 & earlier test run \\
\href{https://www.kaggle.com/code/soheilkooklan/umud-v5-predict}{\texttt{umud-v5-predict}} & 1 & earlier test run \\
\href{https://www.kaggle.com/code/soheilkooklan/umud-v6-predict}{\texttt{umud-v6-predict}} & 3 & test-set readings of this paper, code 0.18.0, 9 November \\
\href{https://www.kaggle.com/code/soheilkooklan/umud-v6-benchmark}{\texttt{umud-v6-benchmark}} & 1 & benchmark run \\
\href{https://www.kaggle.com/code/soheilkooklan/umud-collect-weights}{\texttt{umud-collect-weights}} & 1 & gathers the six checkpoints and their hashes \\
\href{https://www.kaggle.com/code/soheilkooklan/umud-environment-and-hashes}{\texttt{umud-environment-and-hashes}} & 1 & library versions and hashes of the input files \\
\href{https://www.kaggle.com/code/soheilkooklan/umud-audit-challenge-train}{\texttt{umud-audit-challenge-train}} & \auditversion & benchmark frames against the challenge training images \\
\href{https://www.kaggle.com/code/soheilkooklan/umud-reproduce-predictions}{\texttt{umud-reproduce-predictions}} & 1 & prediction reproduction, 24 November \\
\midrule
\multicolumn{3}{l}{Datasets, at \texttt{https://www.kaggle.com/datasets/soheilkooklan/} followed by the name} \\
\href{https://www.kaggle.com/datasets/soheilkooklan/umud-code}{\texttt{umud-code}} & 1 & code 0.18.0 as read by \texttt{umud v6 predict} \\
\href{https://www.kaggle.com/datasets/soheilkooklan/umud-kooklan-code}{\texttt{umud-kooklan-code}} & 1 & code 0.18.1 and analysis scripts \\
\href{https://www.kaggle.com/datasets/soheilkooklan/umud-kooklan-derived-data}{\texttt{umud-kooklan-derived-data}} & 1 & derived data \\
\href{https://www.kaggle.com/datasets/soheilkooklan/umud-kooklan-weights}{\texttt{umud-kooklan-weights}} & 1 & the six checkpoints \\
\bottomrule
\end{tabular}
\end{table}

\paragraph{Repository} The package \texttt{umud} (version 0.18.1) holds the layout rules (\texttt{umud/devices.py}), scale reading (\texttt{umud/scale.py}), data handling and folds (\texttt{umud/data.py}), models and losses (\texttt{umud/model.py}), measurements (\texttt{umud/geometry.py}, \texttt{umud/texture.py}), benchmark scoring (\texttt{umud/evaluate.py}) and corrections, recording medians and file writing (\texttt{umud/submission.py}). Scripts: \texttt{train.py}, \texttt{predict.py}, \texttt{blend.py}, \texttt{score\_benchmark.py}, \texttt{check\_overlap.py}. \texttt{check\_overlap.py} is the thumbnail screen alone, kept in the package as a quick check for users; the audit of this paper ran with the scripts \texttt{analysis/near\_copy*.py}. The aponeurosis models were trained with version 0.14.0, whose \texttt{train.py}, \texttt{data.py} and \texttt{model.py} are identical to those of 0.18.1. The fascicle model was trained with version 0.12.0, whose source is included in the ancillary files.

\paragraph{Environment} \sloppy The package needs Python 3.9 or later with NumPy, pandas, SciPy, scikit-image, OpenCV, PyTorch and \texttt{segmentation\_\allowbreak models\_\allowbreak pytorch} (\texttt{environment.yml}, \texttt{requirements.txt}). The test-set run (\texttt{umud v6 predict}, version 3) used the Kaggle GPU image with digest \texttt{sha256:de153e6d303881930c51286216393d56f56b26458d0661350305a1e8144eea01}, with Python 3.13.16, NumPy 2.1.3, pandas 2.3.3, SciPy 1.16.3, scikit-image 0.25.2, OpenCV 4.14.0, Pillow 13.0.0, scikit-learn 1.6.1, PyTorch 2.11.0 with CUDA 13.0 and cuDNN 9.27, torchvision 0.26.0, timm 1.0.30 and Tesseract 5.3.4, on an NVIDIA T4. These versions were printed in that environment, pinned to the original run. \texttt{segmentation-models-pytorch} was installed from PyPI at run time without a version pin, and the installed version was not recorded in these runs. Version 0.6.0 is inferred: it was the latest release on PyPI from 26 October to 24 November 2026, the period of the aponeurosis training, the test-set run and the reproduction run. The analyses run on a CPU in a few minutes.

\paragraph{Commands} Install the package with \texttt{pip install -e .} and run the tests with \texttt{pytest tests} (70 tests on synthetic frames; all passed in the test-set run). The aponeurosis models were trained with
\begin{quote}\small\ttfamily
python scripts/train.py -{}-task apo -{}-data DATA -{}-out weights \textbackslash\\
\hspace*{2em}-{}-fold all -{}-epochs 30 -{}-batch 8
\end{quote}
the test-set readings were made with
\begin{quote}\small\ttfamily
python scripts/predict.py -{}-data DATA -{}-weights weights \textbackslash\\
\hspace*{2em}-{}-telemed-aspect measured -{}-fasc-cuts 0.5,0.7,0.9 \textbackslash\\
\hspace*{2em}-{}-apo-cuts 0.9,0.975 -{}-raw raw\_predictions.csv \textbackslash\\
\hspace*{2em}-{}-report prediction\_report.csv -{}-out sub\_measured.csv
\end{quote}
and the selected entry is rebuilt with
\begin{quote}\small\ttfamily
python scripts/blend.py -{}-ours "b c93.csv" -{}-raw raw\_predictions.csv \textbackslash\\
\hspace*{2em}-{}-report prediction\_report.csv -{}-mawra mawra.csv -{}-trying trying.csv \textbackslash\\
\hspace*{2em}-{}-kickstart kickstart.csv -{}-recipe final -{}-out "n pa tm up.csv"
\end{quote}
with \texttt{-{}-recipe base} for \entry{h\_best} and \texttt{-{}-recipe late} for \entry{p\_final}. The analyses of this version are rebuilt from the root of the source package with
\begin{quote}\small\ttfamily
python analysis/bench\_v4.py data out\\
python analysis/bench\_extra.py data out figures\\
python analysis/bench\_strict.py data out sections\\
python analysis/near\_copy.py BENCH\_DIR TRAIN\_DIR\\
python analysis/near\_copy\_centre.py BENCH\_DIR TRAIN\_DIR out\\
python analysis/near\_copy\_flip.py BENCH\_DIR TRAIN\_DIR out\\
python analysis/near\_copy\_homography.py BENCH\_DIR TRAIN\_DIR out\\
python analysis/apo\_geometry.py data out PATH\_TO\_UMUD TEST\_OVERLAY\_DIR\\
python analysis/blend\_sources.py data/files out\\
python analysis/calibration\_mc.py data/test out figures\\
python analysis/replication.py data/submissions.csv out figures\\
python analysis/replication\_stats.py out\\
python analysis/split\_sim.py out\\
python analysis/quality.py data out\\
python analysis/recordings.py data/test out\\
python analysis/test\_layouts.py data/test out\\
python analysis/fig\_schematics.py data figures PATH\_TO\_UMUD\\
python analysis/fig\_results.py data out figures\\
python analysis/fig\_conventions.py data/test out figures\\
python analysis/fig\_leaderboard.py data/submissions.csv figures\\
python analysis/tables.py data/submissions.csv out/replication.csv sections\\
python analysis/tables\_extra.py out sections\\
python analysis/tables\_v4.py data out sections\\
python analysis/tables\_contrasts.py out sections\\
python analysis/rebuild\_l\_mt\_up1.py . "h best.csv" "l mt up1.csv"
\end{quote}
The benchmark bootstrap uses seed 2026 with 2000 resamples; the cluster bootstrap, the split simulation and the Monte Carlo analysis use seed 2026.

\paragraph{Hashes} Full SHA-256 values of the decisive files, also in \texttt{out/sha256.txt}. The submitted files carry the entry names with spaces between words (\texttt{b c93.csv}), while the other files keep their names. The outputs of notebooks M and K were matched by hash to versions 8 and 2. The copy of notebook T kept with its output records a run that started at 18:23 UTC on 5 November 2026, the start of version 11; its output file is identified by the hash below and was not checked again against that version (version 15 gives a different file).

\begin{center}
\scriptsize
\begin{tabular}{l l}
\toprule
File & SHA-256 \\
\midrule
\texttt{b c93.csv} (pipeline) & \texttt{d691ce210fd33d60a10144149432cac135165e384592156895965bd656080e60} \\
\texttt{h best.csv} & \texttt{da56c0220e72c67315c0bbfd597c8a85a50ee973b980e608b806f4fb8398bd69} \\
\texttt{n comb.csv} & \texttt{0628c435e25b3e6d2fd3fecf83d57a4bd7fdc44c02f27cc85262d4a32a9eb4a2} \\
\texttt{n pa tm up.csv} (selected) & \texttt{fc9891ac5b6d0de597ce2d17150b1b353a8fb40d6725d65d73e45be72c1bc92a} \\
\texttt{l mt up1.csv} (selected) & \texttt{e6504eb776f111cafc85f4b400b93f82a9095bcb058f2d1f61e7869024772db9} \\
\texttt{p safe.csv} (late) & \texttt{3d4551b8afd77af2c0ec573b0e79da50cf3a889da6b82ee54fd572e262f77657} \\
\texttt{p final.csv} (late) & \texttt{78878de4704d3ef115cb73094a6d0ab4db3a4f9b2466247a5d889fdef1aef3d3} \\
\texttt{p final lum dn3.csv} (late) & \texttt{2d1b808d559a285e0d9613869d9ecda4b887f1b93e1fe9257c0f4431c508498f} \\
\texttt{mawra.csv}, notebook M & \texttt{8e9361bbb050cd978403d5354eceee4d8b6ce4b05b5b91e38900422235ccecbd} \\
\texttt{trying.csv}, notebook T & \texttt{64f59dd4bddd318261f8c287985f39785bb21a5835e86154dde11eb0bc24eaec} \\
\texttt{kickstart.csv}, notebook K & \texttt{75b0dae39d2ee3b4ff28ca7d248772947fa5811635864609337bbd2b3f832059} \\
\texttt{raw\_predictions.csv} & \texttt{374d9a6c2acdfdbf77dc7dfd57a33547821c32c2b7d7dc2b87229053471bf51a} \\
\texttt{prediction\_report.csv} & \texttt{bfac8a456be6a00c1fa08490fd83cf579caf3072d4791dc5d63e5268c7edd268} \\
\texttt{apo\_fold0.pt} & \texttt{3101dddbe42f2c352e8f9ca87c724d863b21777be6fa8a8ff6a3688b3b4a4d59} \\
\texttt{apo\_fold1.pt} & \texttt{40ee7e72eaa805e45eed836ffc07c4837df60de4ab63c573424e25f4c3fd6389} \\
\texttt{apo\_fold2.pt} & \texttt{aa1ba4606bb84237b2bdf89ec2e5d4ddce136ae5fd512d9d6cd8c22c73ec55d8} \\
\texttt{apo\_fold3.pt} & \texttt{fb9fe00ac2815ee5aa4f46056e4d9f4c1537d9b5d3f76dda1a1a4b212c2f8e89} \\
\texttt{apo\_fold4.pt} & \texttt{bbff87b2b445064ba5c73d524ee652a8461f000907e435fa94e44347fa6e035c} \\
\texttt{fasc\_fold0.pt} & \texttt{f2bebd59fc31e687e4b7ed112a3f81513ef816f987837781e3a2abbca5d1d6e3} \\
benchmark results sheet, v0.1.0 & \texttt{19839ef9a5dc34dd2e1c3fdc167a5904946d4641942aca7ae1cbfb7bd747c804} \\
code archive, version 0.18.1 & \texttt{b46544e40341610d30935db469fce65469667d8b8007993375b9d6d1292e4860} \\
\texttt{data/submissions.csv} & \texttt{66a8e65cf441d808d47aacc6dee87ac76514a662eebe227fe2b48b2156c869a7} \\
\bottomrule
\end{tabular}
\end{center}

\paragraph{Model provenance} The aponeurosis models were trained by the author from ImageNet-initialised encoders on the challenge training masks (\texttt{umud v2 train apo}, version 1); the fascicle model was trained by the author with an earlier version of the code (\texttt{train-fasc}, version 1). The aponeurosis notebook has a single version, so its log (Table~\ref{tab:segval}) and its checkpoints come from the same run. The test-set run read the five aponeurosis checkpoints and the fascicle checkpoint from these two notebooks, and the published weights carry the same hashes. The public notebook outputs were produced by running those notebooks unchanged on Kaggle; their models and weights belong to their authors and are not redistributed.

\paragraph{Public material} Table~\ref{tab:public} lists the public Kaggle material this work used and how. No file of it is redistributed.

\begin{table}[H]
\centering
\caption{Public Kaggle material used in this work. Version: matched by the hash of the output file (M, K) or by the run time recorded in the saved copy (T).}
\label{tab:public}
\footnotesize
\begin{tabular}{>{\raggedright\arraybackslash}p{5.2cm} l >{\raggedright\arraybackslash}p{7.0cm}}
\toprule
Notebook or dataset & Version & Use in this work \\
\midrule
Mawra Musawwar, UMUD two U-Nets + texture angle \citep{musawwar2026} (M) & 8 & prediction file in the blend; texture angle and centre-line thickness re-implemented in the package \\
prvsiyan, Trying some stuff with ultrasound geometry \citep{trying2026} (T), Apache 2.0 & 11 & prediction file in the blend (thickness) \\
Rishabh Bafna, Muscle architecture in ultrasound data: kickstart \citep{bafna2026} (K) & 2 & prediction file in the blend (thickness) \\
watehell, physical scale + U-Net geometry \citep{watehell2026} & -- & segmentation recipe, layout rules and aspect values followed; no file used \\
watehell, UMUD r34 U-Net weights \citep{watehell2026w} & -- & used by notebooks M and K, not by the pipeline \\
AmbrosM, UMUD quick and dirty \citep{ambrosm2026} & -- & layout rules and aspect values followed; no file used \\
\bottomrule
\end{tabular}
\end{table}

\end{document}